%% file: main.tex
\documentclass[runningheads]{llncs}

\usepackage{eccv}

\usepackage{eccvabbrv}

\usepackage{graphicx}
\usepackage{booktabs}
\usepackage[table]{xcolor}

\usepackage[accsupp]{axessibility}  

\usepackage{booktabs}
\usepackage{array}
\usepackage{tabularx}
\usepackage{stfloats}
\usepackage{caption}

\usepackage[hidelinks,breaklinks,colorlinks,citecolor=eccvblue]{hyperref}

\usepackage{orcidlink}
\usepackage{bibunits}

\usepackage{enumitem}

\makeatletter
\let\tf@llncs@section\section
\let\tf@llncs@subsection\subsection
\let\tf@llncs@subsubsection\subsubsection
\let\tf@llncs@subparagraph\subparagraph
\let\subparagraph\paragraph
\makeatother
\usepackage{titlesec}
\makeatletter
\let\subparagraph\tf@llncs@subparagraph
\makeatother
\titlespacing*{\section}{0pt}{6pt plus 0pt minus 0pt}{6pt plus 0pt minus 0pt}
\titlespacing*{\subsection}{0pt}{4pt plus 0pt minus 1pt}{4pt plus 0pt minus 1pt}
\titlespacing*{\subsubsection}{0pt}{3pt plus 0pt minus 1pt}{3pt plus 0pt minus 1pt}

\begin{document}

\setlist[itemize]{label=\textbullet, topsep=0.3\baselineskip, itemsep=0.15\baselineskip, parsep=0pt, partopsep=0pt}

\setlength{\abovedisplayskip}{3pt plus 0.5pt minus 0.5pt}
\setlength{\belowdisplayskip}{3pt plus 0.5pt minus 0.5pt}
\setlength{\abovedisplayshortskip}{0.5pt plus 0.5pt}
\setlength{\belowdisplayshortskip}{0.5pt plus 0.5pt}

\title{Trajectory Forcing: Structure-First Generation with Controllable Semantic Trajectories
}

\titlerunning{Trajectory Forcing}

\author{
Merve Kocabas\inst{1,3}\orcidlink{0009-0005-6310-5701} \and
Gege Gao\inst{1,2}\orcidlink{0000-0002-6918-2928}\textsuperscript{\textdagger} \and \\
Bernhard Sch\"olkopf\inst{2,3,4,5}\orcidlink{0000-0002-8177-0925}\textsuperscript{\textdaggerdbl} \and
Andreas Geiger\inst{1,4,5}\orcidlink{0000-0002-8151-3726}\textsuperscript{\textdaggerdbl}
}

\authorrunning{M. Kocabas et al.}

\institute{
University of T\"ubingen, T\"ubingen, Germany
\email{merve.kocabas@student.uni-tuebingen.de}\\[-0.2em]
\email{\{gege.gao,a.geiger\}@uni-tuebingen.de}
\and
ETH Z\"urich, Z\"urich, Switzerland 
\and
Max Planck Institute for Intelligent Systems, T\"ubingen, Germany 
\email{bs@tuebingen.mpg.de}
\and
ELLIS Institute, T\"ubingen, Germany \and
T\"ubingen AI Center, T\"ubingen, Germany
\\[0.4em] 
\textsuperscript{\textdagger} Project lead. 
\quad 
\textsuperscript{\textdaggerdbl} Shared last authorship. 
\\[0.4em]
\url{https://mervekocabas.github.io/TrajectoryForcing/}}

\maketitle

\vspace{-3mm}
\begin{abstract}
Diffusion and flow-based generative models produce strong images, yet their controllability remains largely endpoint-centric: users specify conditions and receive final outputs, while the intermediate generative dynamics remain hidden. Recent methods have begun to exploit generation order and process decomposition to improve sample quality, but still treat intermediate states as internal computation rather than objects for interaction. We propose \textbf{Trajectory Forcing (TF)}, a trajectory-centric framework that makes the generation path explicit, semantic, and editable. TF organizes synthesis as a sequence of semantically structured stages, progressing from global layout to object-, part-, and detail-level representations. Each stage produces a decodable latent state that can be inspected, evaluated, and locally edited before the next stage begins. 
To instantiate this path, we derive coarse-to-fine teacher hierarchies by clustering pretrained visual representations such as DINOv2, and train a hierarchy-conditioned one-step flow-matching model at each level.
We further introduce trajectory-aware metrics that measure structural consistency and local controllability beyond endpoint quality metrics such as FID. Experiments show that TF achieves competitive sample quality while exposing coherent intermediate states and supporting localized edits across semantic levels. By shifting the focus from final images to the generative path itself, TF opens a route toward controllable, trajectory-aware image synthesis.

\end{abstract}

\vspace{-3mm}

\input{sec/intro}

\input{sec/rel}

\input{sec/back}

\input{sec/method}

\input{sec/exp}

\section{Discussion and Conclusions}
We presented Trajectory Forcing, an approach that treats the generative trajectory as a first-class, user-facing object rather than a hidden computational process.
By operating in a semantically structured representation space and combining hierarchical conditioning with one-step flow matching, TF produces a coarse-to-fine sequence of decodable, editable intermediate states in $L{=}4$ NFE.

\noindent \textbf{Limitations.}
Our hierarchy is derived from unsupervised clustering with a fixed depth and a center prior for object/background separation, which may not generalize well to images without a centered dominant object.
Because each level is conditioned only on the immediately preceding one, coarser-level information reaches finer levels indirectly through the chain rather than via direct conditioning.
Current results are reported at 40/80 training epochs; longer training and scaling analysis are deferred to the supplementary material.

\noindent \textbf{Future work.}
Natural extensions include replacing the fixed clustering pipeline with richer hierarchy sources (such as model segmentations~\cite{carion2025sam3}), adaptive hierarchy depth per image, and extending TF to text-conditional generation.

\clearpage
\clearpage
\clearpage

\section*{Acknowledgements}
Merve Kocabas is supported by the Konrad Zuse School of Excellence in Learning and Intelligent Systems (ELIZA) through the DAAD programme Konrad Zuse Schools of Excellence in Artificial Intelligence, sponsored by the German Federal Ministry of Education and Research. 
Gege Gao acknowledges the EuroHPC Joint Undertaking for awarding access to computational resources under project ID EHPC-AIF-2026PG01-147 on the Discoverer+ GPU partition hosted by Sofia Tech Park.
Andreas Geiger is a member of the Machine Learning Cluster of Excellence, EXC 2064/1, project number 390727645. 
The authors gratefully acknowledge the ML Cloud and the Tübingen AI Center for providing the computational resources and facilities that supported this work.

%
%
\bibliographystyle{splncs04}
\bibliography{bibliography_long,egbib}

\clearpage
\begingroup
\makeatletter
\let\section\tf@llncs@section
\let\subsection\tf@llncs@subsection
\let\subsubsection\tf@llncs@subsubsection
\makeatother
\input{appendix}

\endgroup
\end{document}

%% file: sec/intro.tex
\begingroup
\raggedbottom

\begin{figure*}
    \centering
    \includegraphics[width=0.8\linewidth]{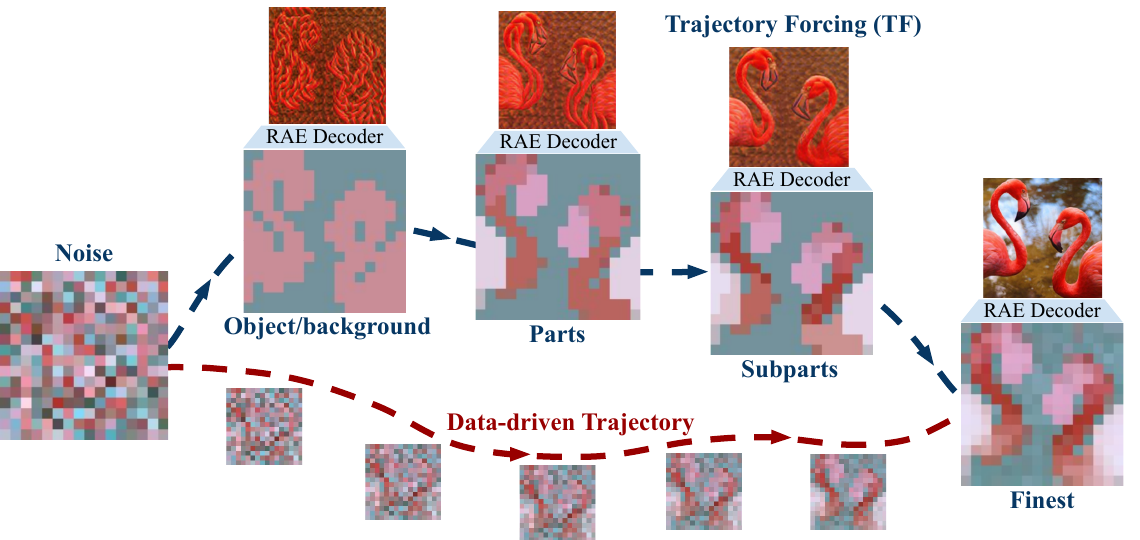}
    \vspace{-2mm}
    \captionsetup{font={stretch=0.8,footnotesize}}
    \captionof{figure}{
        \textbf{Trajectory Forcing (TF)} replaces the opaque, data-driven denoising trajectory (bottom, red) with a deliberately structured coarse-to-fine progression through semantic levels (top, blue). Every intermediate state is decodable via a shared decoder into a visually interpretable image, enabling inspection and editing at each level.
    }
    \label{fig:teaser}
    \vspace{-5mm}
\end{figure*}

\section{Introduction}
\label{sec:intro}

\begin{sloppypar}

\begin{center}
\emph{To control generation, we must expose not only its destination, but also its path.}
\end{center}

Modern image generators typically cast synthesis as a direct mapping from noise to a finished image. Although denoising traverses intermediate states, these states are not deliberately structured for human understanding or intervention; they are computational by-products discarded after the final output is produced. This contrasts with human visual creation, which is inherently coarse-to-fine: global composition is established before local details are refined, and each intermediate stage can be inspected and revised. We take this contrast as our starting point and ask whether generative trajectories themselves can be made interpretable, semantically structured, and editable.

Recent works have begun to exploit trajectory structure in generation~\cite{chen2025diffusionforcing,wang2026nvg,ma2025deco,baade2026latentforcing}, showing that generation order and process decomposition can substantially affect output quality. However, these trajectories remain optimized for the final image: intermediate states are used as computational scratchpads, frequency components, or token-completion steps, rather than as semantic objects a user can inspect, compare, and redirect.

We therefore propose \textbf{Trajectory Forcing (TF)}, a framework that organizes image synthesis as a sequence of semantically structured stages. Each stage produces an interpretable latent state that can be decoded, evaluated, and edited before the next stage begins. Rather than collapsing generation into a single opaque mapping from noise to image, TF exposes a coarse-to-fine hierarchy from global layout, through object-level structure, to fine-grained detail, making every intermediate level visually meaningful and amenable to controlled intervention.

Making this hierarchy operational requires a representation space whose geometry organizes semantic structure. Raw pixels entangle appearance, layout, and identity, whereas pretrained visual representations provide a more suitable substrate. Following the \emph{representation alignment hypothesis}~\cite{yu2025repa}, we operate in DINOv2 feature space~\cite{oquab2023dinov2}, where hierarchical clustering over latent tokens empirically recovers object-part-subpart decompositions (Sec.~\ref{sec:teacher}) that supervise our trajectory design.

A natural concern with multi-stage generation is inference cost: progressive formulations typically require network function evaluations (NFE) that scale as $L\times T$, where $L$ is the number of structural stages and $T$ the number of denoising steps per stage. TF addresses this by integrating one-step flow matching~\cite{geng2025mf,geng2025imf,lu2026pmf} at each stage. Instead of iterative denoising at every level, each stage uses a single function evaluation to map noise to the current-level target, conditioned on the previous stage. This reduces inference to $L$-NFE, making TF comparable to modern few-step generators while preserving trajectory-level controllability.

Our contributions are as follows:
\begin{itemize}
    \item We propose \textbf{Trajectory Forcing (TF)}, a generation framework that models image synthesis as a structured coarse-to-fine trajectory of interpretable semantic stages, enabling inspection and editing at every level.
    \item We introduce a \textbf{hierarchy-conditioned one-step flow} formulation that extends one-step generation to multi-level trajectories via level-wise conditioning and structural consistency supervision.
    \item We design \textbf{trajectory-aware metrics} that go beyond FID to measure structural consistency and local controllability across generation levels.
    \item We validate TF on class-conditional ImageNet generation, achieving competitive image quality and fast FID convergence while enabling interpretable multi-level trajectories and interactive editing.
\end{itemize}

\end{sloppypar}
\endgroup

%% file: sec/rel.tex
\begingroup
\raggedbottom
\section{Related Work}
\label{sec:related}
\begin{sloppypar}

\subsection{Diffusion and Flow-Based Generation}

\noindent \textbf{Latent-space models.}
Latent Diffusion Models~\cite{rombach2022high} established the paradigm of performing generative dynamics in a compressed latent space, decoupling representation learning from the denoising process.
Subsequent work scales the backbone with Transformers~\cite{peebles2023scalable,ma2024sit} or revisits the representation interface through alternative autoencoding schemes~\cite{zheng2026rae}.

\noindent \textbf{Pixel-space models.}
A parallel line of work revisits pixel-space generation, removing reliance on pretrained tokenizers.
Recent approaches include flow-based pixel modeling~\cite{chen2025pixelflow}, neural-field diffusion~\cite{wang2026pixnerd}, and end-to-end training with self-supervised pretraining~\cite{lei2026novae}. 
JiT~\cite{li2026jit} shows that direct clean-image prediction enables diffusion in high-dimensional pixel spaces, establishing a strong pixel-space baseline. 
While these works show that compression-based latents are not strictly necessary, our use of a representation space is motivated by semantic structure rather than compression: pretrained features support both hierarchy construction and visual decodability of intermediate states.

\noindent \textbf{One and few-step models.}
Reducing inference cost is a major research direction~\cite{wang2025tim,zhang2025alphaflow,hu2025mfrae}.
Distillation-based methods compress multi-step models~\cite{salimans2022progressive,meng2023distillation}.
Consistency models learn to map arbitrary intermediate states to the ODE endpoint~\cite{song2023consistency,song2024improved,lu2025simplifying}, with trajectory-aware extensions~\cite{kim2024ctm}.
More recently, shortcut parameterizations~\cite{frans2025one,zhou2025imm} and reparameterized matching objectives, including Mean Flows~\cite{geng2025mf,geng2025imf,lu2026pmf} and drifting-based formulations~\cite{deng2026drifting}, enable one-step or few-step generation.
Our work inherits the mean-flow framework 
and extends it with hierarchical conditioning and structural supervision.

\subsection{Generation Ordering and Trajectory Design}

The ordering of generation has recently emerged as an important design axis for controlling \textit{how information is revealed} during synthesis.

\noindent \textbf{Per-token noise scheduling.}
Diffusion Forcing~\cite{chen2025diffusionforcing} assigns independent noise levels to individual tokens, enabling flexible generation orderings in sequential data.
This idea is theoretically grounded: different conditioning orders can lead to exponential differences in the learnability of data distributions~\cite{diffusionintractable}.

\noindent \textbf{Frequency-domain reordering.}
DeCo~\cite{ma2025deco} factorizes pixel diffusion along frequency components, using a lightweight decoder for high-frequency details while the main DiT specializes in low-frequency semantics.
This decoupling improves both training efficiency and generation quality for pixel-space models.

\noindent \textbf{Latent-pixel reordering.}
Latent Forcing~\cite{baade2026latentforcing} jointly diffuses self-supervised latent features and pixels with separate time schedules.
By revealing latent structure before pixel detail, it achieves the convergence benefits of latent diffusion while remaining end-to-end in pixel space.
The generated latent features serve as a computational scratchpad and are discarded after generation.

All of these methods use trajectory structure to improve \emph{generation quality}, while intermediate states remain either implicit or disposable.
In contrast, our work treats the structured trajectory as a user-facing interface: intermediate stages are designed to be semantically interpretable, decodable, and editable, enabling trajectory-level control that goes beyond endpoint optimization.

\subsection{Progressive and Hierarchical Generation}

\begin{figure*}[t]
    \centering
    \includegraphics[width=0.98\linewidth]{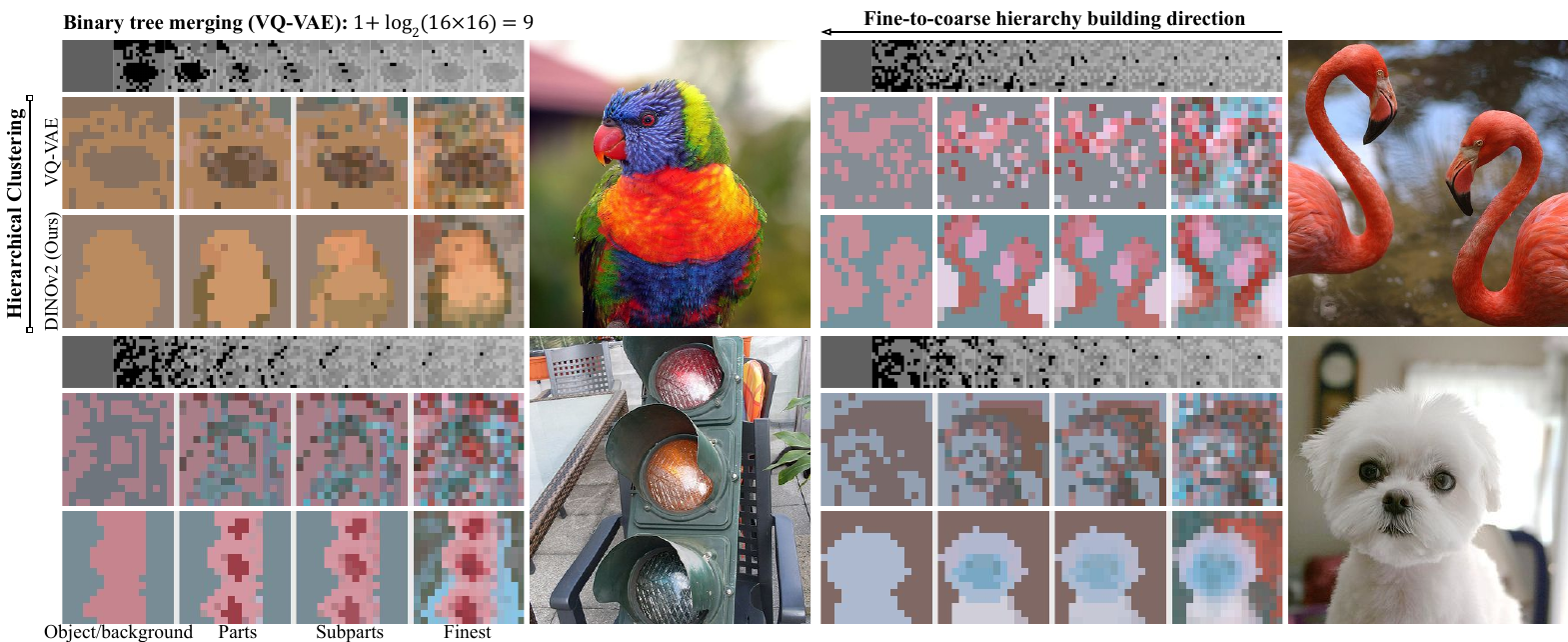}
    \captionsetup{hypcap=false}
    \captionof{figure}{
        \textbf{Hierarchy comparison.}
        NVG's binary merging in VQ-VAE latent space produces a resolution-tied hierarchy ($1+\log_2(16{\times}16)=9$ levels) whose intermediate stages lack clear semantic groupings.
        In contrast, clustering DINOv2 features recovers a compact object-part-subpart hierarchy with semantically interpretable levels.
    }
    \label{fig:hierarchy}
    \vspace{-5mm}
\end{figure*}

\noindent \textbf{Autoregressive and masked models.}
Token-based approaches generate images by predicting discrete tokens~\cite{esser2021taming} or masked subsets~\cite{li2024mar}, sometimes combined with diffusion-based training~\cite{gu2024diffusion,li2024darl}.
These naturally expose partial intermediate samples, but progression is defined over token completion rather than continuous semantic refinement.

\noindent \textbf{Scale-level progression.}
Visual Autoregressive Modeling (VAR)~\cite{tian2024var} and FlowAR~\cite{ren2024flowar} organize generation along spatial resolution, predicting coarse structures first and refining at higher resolutions.
Our hierarchy is orthogonal: levels correspond to semantic granularity (object, part, subpart) at a fixed spatial resolution, rather than resolution upscaling.

\noindent \textbf{Granularity-level progression.}
Explicit semantic or object-level structure has been used to organize visual scenes in related settings~\cite{gao2024graphdreamer,gao2026loopowm,gao2021causal,venkataramanan2025franca,videoflextok}. Most closely related to our hierarchical design, Next Visual Granularity (NVG)~\cite{wang2026nvg} decomposes images into granularity levels by bottom-up clustering in VQ-VAE latent space, and autoregressively predicts structure maps and tokens at each level.
While NVG demonstrates the value of explicit structure control, its hierarchy is constrained by greedy binary L2 merging ($k=2$) in VQ-VAE space, whose limited semantic geometry makes meaningful groupings, such as object-part-subpart, difficult.
Consequently, its stages are tied to latent resolution (e.g., $\log_2(16{\times}16)=8$), its intermediates lack clear semantic abstractions (Fig.~\ref{fig:hierarchy}), and its discrete codebook limits intermediate-state flexibility and continuity.

Our approach shares NVG's goal of making structure explicit in generation, but differs in substrate, dynamics, and interface: we build semantically grounded hierarchies by clustering geometrically regular pretrained representations (e.g., DINOv2~\cite{oquab2023dinov2}), use one-step flow matching per level for $L$-NFE inference instead of $L{\times}T$, and keep all intermediate states continuously decodable and editable for interactive trajectory-level control.

\subsection{Representation-Space Generation}

Recent work has shown that pretrained visual representations can improve generative modeling.
REPA~\cite{yu2025repa} aligns intermediate diffusion features with DINOv2~\cite{oquab2023dinov2} latents via an auxiliary loss, accelerating convergence.
Representation Autoencoders (RAE)~\cite{zheng2026rae} further train decoders to reconstruct images from DINOv2 features, enabling generation in a semantically structured and visually decodable representation space.
This provides the two ingredients our framework relies on: a geometry suitable for coarse-to-fine hierarchies, and a decoder that makes each intermediate level both semantic and visually interpretable.

\end{sloppypar}
\endgroup

%% file: sec/back.tex
\section{Background}

We build on two technical ingredients: (i)~a flow-matching formulation that enables one-step generation, and (ii)~a pretrained representation space whose geometric structure supports both meaningful hierarchical decomposition and visual decoding of intermediate states.

\subsection{Flow Matching and Mean Flows}
We briefly review the progression from flow matching to one-step mean-flow generators, culminating in the backbone we build upon.

\noindent \textbf{Flow Matching.}
Flow Matching~\cite{liu2022flow} learns a velocity field that transports between a prior distribution and the data distribution.
Given data $\mathbf{x} \sim p_{\text{data}}$ and noise $\boldsymbol{\epsilon} \sim \mathcal{N}(\mathbf{0}, \mathbf{I})$, a flow path is defined by the linear interpolation 
\begin{equation}
\mathbf{z}_t = (1{-}t)\,\mathbf{x} + t\,\boldsymbol{\epsilon},
\end{equation} 
with $t \in [0,1]$.
The conditional velocity is $\mathbf{v} = \boldsymbol{\epsilon} - \mathbf{x}$, and a network $\mathbf{v}_\theta$ is trained by minimizing $\mathcal{L}_{\text{FM}} = \mathbb{E}_{t,\mathbf{x},\boldsymbol{\epsilon}} \|\mathbf{v}_\theta(\mathbf{z}_t, t) - \mathbf{v}\|^2$.
Samples are generated by solving the ODE $\tfrac{d}{dt}\mathbf{z}_t = \mathbf{v}_\theta(\mathbf{z}_t, t)$ from $t{=}1$ (noise) to $t{=}0$ (data), typically requiring many network evaluations.

\noindent \textbf{Mean Flows.}
MeanFlow (MF)~\cite{geng2025mf} enables one-step generation by modeling an \emph{average velocity} field.
Viewing Flow Matching's $\mathbf{v}(\mathbf{z}_t, t)$ as the \emph{instantaneous} velocity, MF defines the average velocity over an interval $[r, t]$ as:
\begin{equation}
    \mathbf{u}(\mathbf{z}_t, r, t) \;\triangleq\; \frac{1}{t - r} \int_{r}^{t} \mathbf{v}(\mathbf{z}_\tau, \tau)\, d\tau.
    \label{eq:avg_vel}
\end{equation}
Since directly evaluating this integral during training is intractable, MF differentiates both sides with respect to $t$ to derive the \emph{MeanFlow Identity}:
\begin{equation}
    \mathbf{u}(\mathbf{z}_t, r, t) = \mathbf{v}(\mathbf{z}_t, t) - (t - r)\,\frac{d}{dt}\mathbf{u}(\mathbf{z}_t, r, t),
    \label{eq:mf_identity}
\end{equation}
where the total derivative $\tfrac{d}{dt}\mathbf{u}$ is computed via a Jacobian-vector product (JVP).
A network $\mathbf{u}_\theta(\mathbf{z}_t, r, t)$ is trained to satisfy this identity using the conditional velocity $\mathbf{v}(\mathbf{z}_t, t)$ as ground-truth signal and a stop-gradient on the JVP term, denoted $\mathrm{JVP}_{\text{sg}}$~\cite{geng2025mf}.
Once trained, one-step sampling reduces to $\mathbf{z}_0 = \mathbf{z}_1 - \mathbf{u}_\theta(\mathbf{z}_1, 0, 1)$ with $\mathbf{z}_1 \sim \mathcal{N}(\mathbf{0}, \mathbf{I})$.

Improved MeanFlow (iMF)~\cite{geng2025imf} reformulates the objective as a $\mathbf{v}$-loss re-parameterized through $\mathbf{u}_\theta$, constructing a \emph{compound} prediction $\mathbf{V}_\theta = \mathbf{u}_\theta + (t{-}r)\cdot\mathrm{JVP}_{\text{sg}}$ that is regressed against a \emph{guided velocity} $\mathbf{v}_g$.
This resolves a variance amplification issue in the original JVP computation.
Furthermore, iMF incorporates classifier-free guidance (CFG) directly into training by constructing the guided velocity as: 
\begin{equation}
\mathbf{v}_g = \omega\,\mathbf{v}(\mathbf{z}_t \mid \mathbf{c}) + (1{-}\omega)\,\mathbf{v}(\mathbf{z}_t),
\end{equation}
where $\omega$ is a guidance scale sampled at training time and provided to the network $\mathbf{u}_\theta$ alongside the class label $y$ and time interval $[r,t]$ as global conditioning.
This enables flexible adjustment of the guidance strength at inference without retraining.
We collectively denote these \emph{standard} conditions (class label, $[r,t]$, and $\omega$) as $\mathbf{c}$ hereafter, distinguishing them from the hierarchy-specific conditions introduced in Sec.~\ref{sec:denoising}.

\noindent \textbf{$\mathbf{x}$-prediction.}
Orthogonal to the training objective, the choice of prediction target also matters.
Pixel MeanFlow (pMF)~\cite{lu2026pmf} introduces a denoised-image field $\hat{\mathbf{x}}(\mathbf{z}_t, r, t) \triangleq \mathbf{z}_t - t \cdot \mathbf{u}(\mathbf{z}_t, r, t)$ and lets the network directly predict the denoised data $\hat{\mathbf{x}}$.
The average velocity is recovered via $\mathbf{u}_\theta = \tfrac{1}{t}(\mathbf{z}_t - \hat{\mathbf{x}}_\theta)$.
This parameterization is motivated by the manifold hypothesis adopted from JiT~\cite{li2026jit}: the denoised output $\hat{\mathbf{x}}$ lies approximately on a low-dimensional data manifold, making it a more tractable regression target than the noisy velocity field~\cite{lu2026pmf}.
This assumption extends naturally to our setting, where $\hat{\mathbf{x}}$ corresponds to DINOv2 features on a structured representation manifold.
We therefore adopt $\mathbf{x}$-prediction combined with $\mathbf{v}$-loss objective as our generative backbone and extend it with hierarchical conditioning in Sec.~\ref{sec:method}.

\subsection{Representation Autoencoders}

Our framework operates in the feature space of a pretrained visual understanding encoder DINOv2~\cite{oquab2023dinov2}.
This choice is motivated by two properties.
\textbf{First}, the \emph{representation alignment hypothesis}~\cite{yu2025repa} posits that such feature spaces are geometrically well-organized: semantic factors of variation (e.g., object identity, part structure, spatial layout) are well-separated, so that simple unsupervised clustering over DINOv2 tokens reliably recovers object-part-subpart decompositions without any supervised annotations.
We leverage this directly to construct the teacher hierarchies that guide our trajectory design (Sec.~\ref{sec:teacher}). 
\textbf{Second}, Representation Autoencoders (RAE)~\cite{zheng2026rae} train a decoder that reconstructs images directly from DINOv2 features, making any point in this space visually decodable.
Because our generation proceeds as a sequence of intermediate refinements in DINOv2 space, the RAE decoder renders every intermediate level as a visual output, enabling inspection and editing before generating the next stage.

%% file: sec/method.tex
\begin{figure*}[t]
\centering
\includegraphics[width=0.85\linewidth]{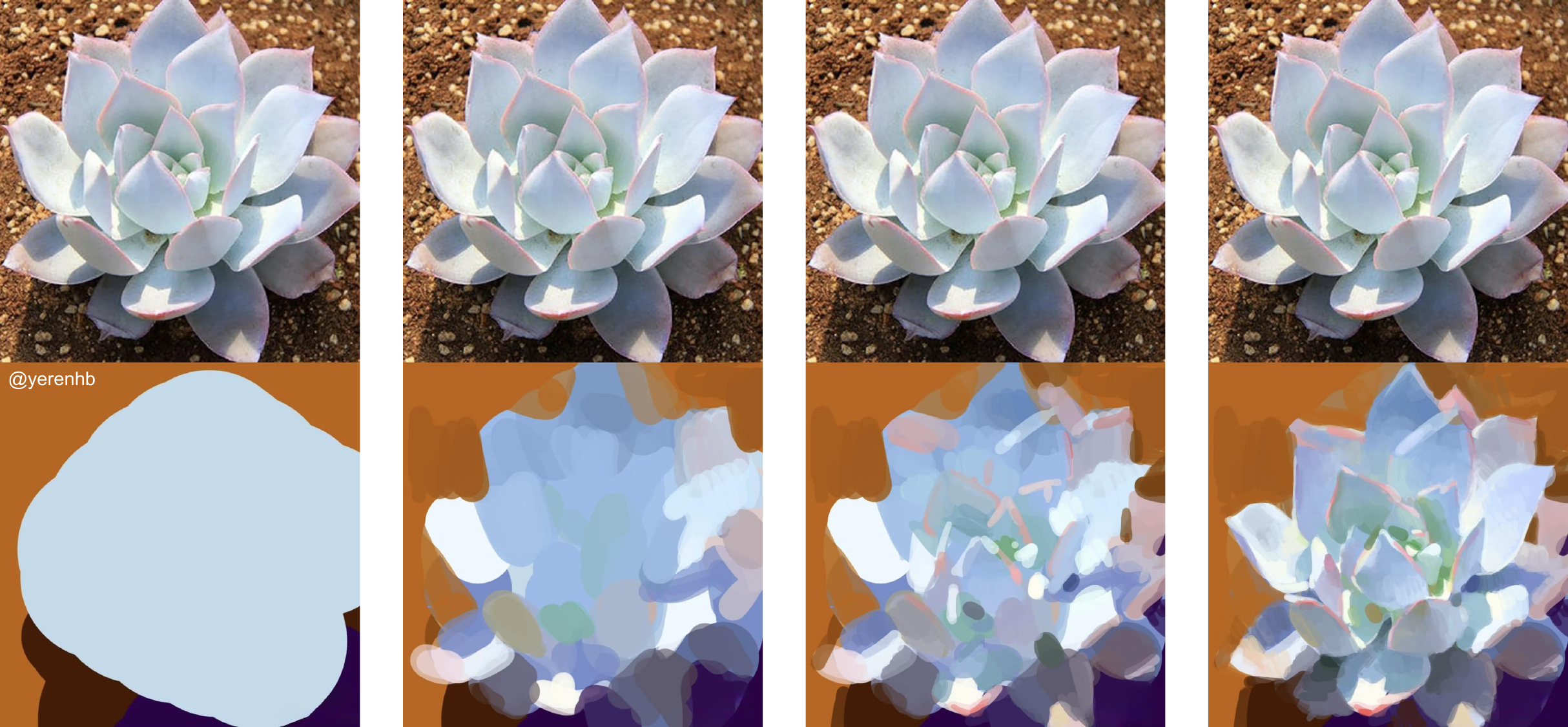}
\captionsetup{font=footnotesize}
\captionof{figure}{
    Speed painting illustrates a coarse-to-fine creation process: artists first establish global shape and color blocks before refining local details, yet intermediate stages are already recognizable. (Images courtesy of @yerenhb.)
}
\label{fig:painting}
\vspace{-5mm}
\end{figure*}

\section{Method}
\label{sec:method}

\noindent\textbf{Artistic training and generative modeling. } Human visual artists typically adopt a coarse-to-fine workflow, first establishing global structure and dominant color relationships before refining local details, as illustrated in Fig.~\ref{fig:painting}. Across artistic training practices, beginners are repeatedly advised to avoid \emph{chasing details} and instead focus on structural abstraction and global coherence. The recurrence of this principle across instructors and contexts suggests that it reflects a stable cognitive regularity: global organization precedes and constrains local articulation.
Motivated by this observation, we incorporate this structural prior into our model design, translating the coarse-to-fine principle into a hierarchical generation framework where global consistency guides detail refinement.

\begin{figure*}[t]
    \centering
    \includegraphics[width=1\linewidth]{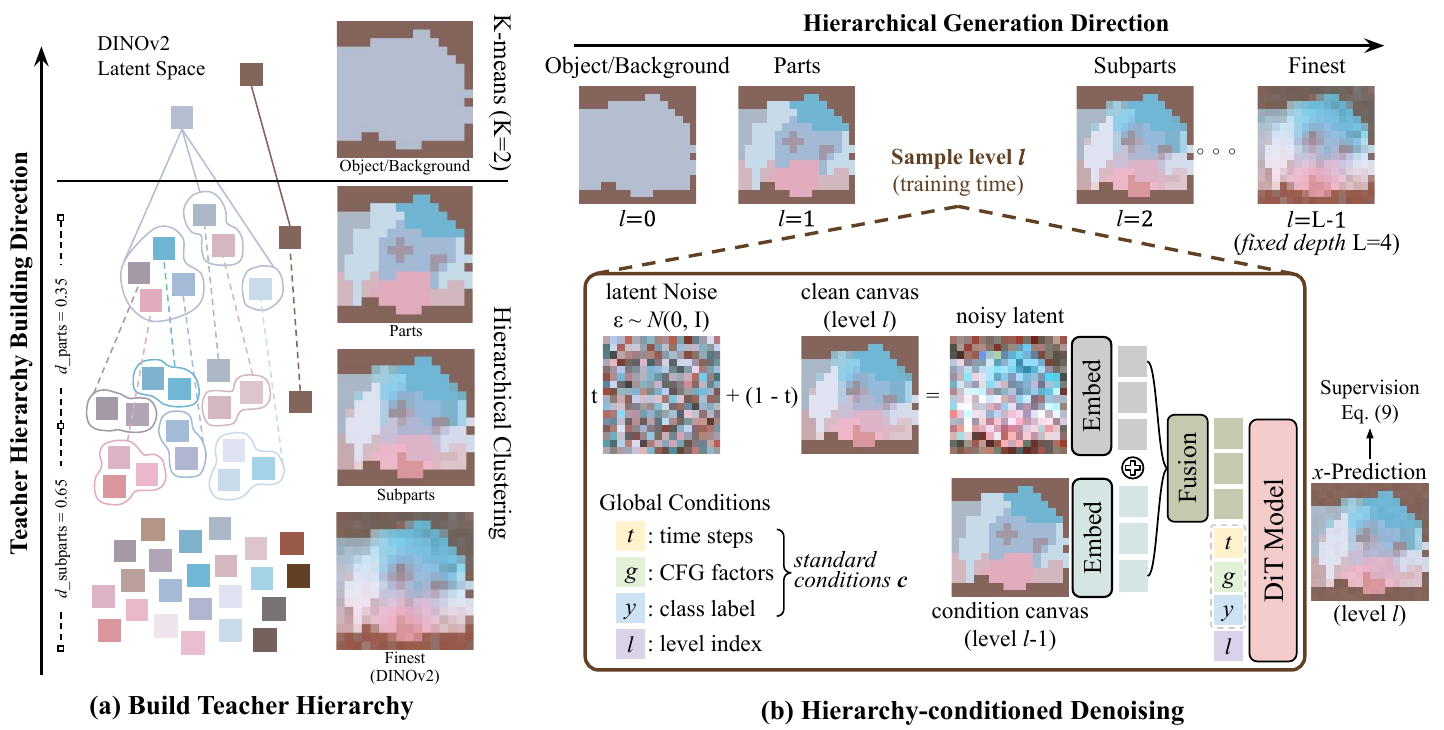}
    \captionsetup{font=footnotesize}
    \vspace{-3mm}
    \captionof{figure}{
        \textbf{Trajectory Forcing pipeline.}
        Given an input image, we extract DINOv2 features and construct a teacher hierarchy via unsupervised clustering, producing level canvases from fine (original features, $l=L-1$) to coarse (object/background, $l=0$).
        A single shared network is trained across all levels: at each training step a level $l$ is sampled, and the network denoises the current-level canvas $\mathbf{z}^{(l)}(t)$ conditioned on the previous-level canvas $\mathbf{z}^{(l-1)}$.
        At inference, the direction reverses: generation proceeds sequentially from coarse to fine. 
        $\oplus$ denotes channel-wise concatenation.
    }
    \label{fig:pipeline}
    \vspace{-7mm}
\end{figure*}

\subsection{Build Teacher Hierarchy}
\label{sec:teacher}

To enable hierarchical generation with intermediate control, we first construct a semantic hierarchy over latent tokens, which serves as a teacher signal for training.
Rather than assuming access to human-annotated part hierarchies, we derive this structure directly from pretrained visual representations, leveraging their inherent semantic organization, as illustrated in Fig.~\ref{fig:pipeline}(a).

Given an input image, we extract dense latent features $\mathbf{z} = \{\mathbf{z}_i\}_{i=1}^{N}$ with $\mathbf{z}_i \in \mathbb{R}^{C}$, where $N{=}H{\times}W$ is the number of spatial tokens.
We assign each token a hierarchical cluster index at \textbf{three granularity levels}: object/background, parts, and subparts, via unsupervised clustering, forming a fixed-depth hierarchy shared across all samples.

\noindent \textbf{(i) Object/background:}
At the coarsest level, we apply K-means with $K{=}2$ in the feature space to obtain two clusters; the cluster whose tokens are on average closer to the image center is designated as object (center prior), and the other as background.
This yields a binary partition that serves as the foundation for finer-grained decomposition.

\noindent \textbf{(ii) Part and Subpart:}
We construct a hierarchy over \textbf{\textit{object tokens only}} via agglomerative clustering with a combined \textit{semantic-spatial distance}:
\begin{equation}
    \mathcal{D}(i,j) = \cos\!\left<\mathbf{z}_i,\mathbf{z}_j\right> + \alpha \, \lVert \mathbf{p}_i - \mathbf{p}_j \rVert_2,
\end{equation}
where $\mathbf{p}_i$ is the normalized spatial coordinate of token $i$ and $\alpha$ controls the spatial regularization strength. 
This encourages semantically coherent, spatially contiguous clusters.
From the resulting dendrogram, we extract a fixed-depth hierarchy by cutting at predefined \emph{distance thresholds}: higher thresholds ($0.65$) yield finer subparts, lower thresholds ($0.35$) produce coarser parts.

\noindent \textbf{(iii) Output Hierarchy:}
The resulting hierarchy assigns each token a tuple of cluster indices (object/background, part, subpart).
These indices are stored as dense maps aligned with the latent grid and serve as region assignments in subsequent stages.

\subsection{Hierarchy-conditioned Denoising}
\label{sec:denoising}

Given the teacher hierarchies from Sec.~\ref{sec:teacher}, we train a hierarchy-conditioned one-step flow model that refines latent features across semantic levels, inducing structured and steerable denoising trajectories.

\subsubsection{Level Targets.}

We construct a set of \emph{level canvases} $\{\mathbf{z}^{(l)}\}_{l=0}^{L-1}$ that serve as denoising targets at each granularity.
For levels $l = 0, \ldots, L{-}2$, each token is replaced by the mean feature of its assigned region:
\begin{equation}
    \mathbf{z}^{(l)}_i = \boldsymbol{\mu}^{(l)}_{R^{(l)}_i}, \qquad
    \boldsymbol{\mu}^{(l)}_k = \frac{1}{|R^{(l)}_k|}\sum_{j \in R^{(l)}_k} \mathbf{z}_j,
    \label{eq:level_canvas}
\end{equation}
where $R^{(l)}_i$ is the region assignment of token $i$ at level $l$.
At the finest $l = L{-}1$, the canvas is the original feature: $\mathbf{z}^{(L-1)} = \mathbf{z}$.
The resulting canvases progress from coarse, piecewise-constant representations to the full-resolution latent, naturally encoding a semantic trajectory from global structure to fine detail.

\subsubsection{Level-wise Conditioning.}

A single shared network is trained across all hierarchy levels.
During training, a level index $l$ is sampled uniformly at random for each batch element.
Beyond the standard conditions $\mathbf{c}$, the network receives three hierarchy-specific inputs: (1)~the noised current-level canvas $\mathbf{z}^{(l)}(t) = (1{-}t)\,\mathbf{z}^{(l)} + t\,\boldsymbol{\epsilon}$, (2)~the previous-level canvas $\mathbf{z}^{(l-1)}$ as a spatial condition, and (3)~the level index $l$ as a global condition.
For level $l{=}0$, the conditioning canvas is set to zero, making the coarsest level unconditional (aside from $\mathbf{c}$).

A natural way to supply the previous-level canvas is \emph{in-context conditioning}~\cite{geng2025imf}: the conditioning tokens are appended to the input token sequence so that the transformer attends over both, but this doubles the sequence length.
For efficiency, we instead fuse the two inputs in the channel dimension: the noisy canvas and the previous-level canvas are independently embedded to $D$ dimensions, concatenated to $2D$, and projected back to $D$ via a linear fusion layer, preserving the original sequence length.
The level index is encoded through a learned embedding table and injected as additional conditioning tokens alongside class and guidance embeddings.

\subsubsection{Total Training Objective.}

\

\noindent \textbf{(i) Flow loss:}
We adopt $\mathbf{x}$-prediction combined with $\mathbf{v}$-loss~\cite{lu2026pmf}.
The network predicts denoised latents $\hat{\mathbf{x}}_\theta(\mathbf{z}^{(l)}(t),\, l,\, \mathbf{z}^{(l-1)};\, \mathbf{c})$,
from which the average velocity is recovered as $\mathbf{u}_\theta = (\mathbf{z}^{(l)}(t) - \hat{\mathbf{x}}_\theta)/t$.
The compound prediction $\mathbf{V}_\theta = \mathbf{u}_\theta + (t{-}r)\cdot\mathrm{JVP}_{\mathrm{sg}}$ is constructed and regressed against the guided velocity $\mathbf{v}_g$:
\begin{equation}
    \mathcal{L}_{\text{flow}} = \mathbb{E}_{t,r,l} \left[\, \| \mathbf{V}_\theta - \mathbf{v}_g \|^2 \,\right].
    \label{eq:flow_loss}
\end{equation}

\noindent \textbf{(ii) Structural loss:}
To encourage the prediction to respect region structure, we penalize the deviation of each predicted token from the ground-truth mean of its assigned region.
Let $\boldsymbol{\mu}^{(l)}_k$ denote the target region mean as in Eq.~(\ref{eq:level_canvas}).
For each token $i$ in region $k$, we compute the squared cosine distance between the predicted token $\hat{\mathbf{x}}_{\theta,i}$ and its region's target mean, averaged over all valid tokens and regions:
\begin{equation}
    \mathcal{L}_{\text{struct}} = \frac{1}{K_l} \sum_{k=1}^{K_l} \frac{1}{|R^{(l)}_k|} \sum_{i \in R^{(l)}_k} \left(1 - \cos\!\left<\hat{\mathbf{x}}_{\theta,i},\, \boldsymbol{\mu}^{(l)}_k\right>\right)^{\!2},
    \label{eq:struct_loss}
\end{equation}
where $K_l$ is the number of valid regions.
This loss is applied to $l \in \{0, \ldots, L-2\}$ and disabled at the finest level, where no region structure is imposed.

\noindent \textbf{Overall objective.}
The full training loss is:
\begin{equation}
    \mathcal{L} = \mathcal{L}_\text{flow} + \lambda \, \mathcal{L}_\text{struct},
    \label{eq:total_loss}
\end{equation}
where $\lambda$ controls the structural loss weight.

\subsection{Hierarchical Sampling}
\label{sec:sampling}

At inference, generation proceeds sequentially through the hierarchy from $l=0$ to $L{-}1$.
At the coarsest level, the model generates from pure noise with zero spatial conditioning:
$\hat{\mathbf{z}}^{(0)} := \hat{\mathbf{x}}_\theta(\boldsymbol{\epsilon},\, l{=}0,\, \mathbf{0};\, \mathbf{c})$.
For each subsequent level $l > 0$, fresh noise is drawn and the model generates conditioned on the output of the previous level:
\begin{equation}
    \hat{\mathbf{z}}^{(l)} := \hat{\mathbf{x}}_\theta(\boldsymbol{\epsilon}',\, l,\, \hat{\mathbf{z}}^{(l-1)};\, \mathbf{c}).
    \label{eq:hier_sample}
\end{equation}
Each level requires a single network function evaluation (NFE), yielding a total cost of $L$-NFE for the complete hierarchy.
Crucially, every intermediate output $\hat{\mathbf{z}}^{(l)}$ is decodable into a visual image via the RAE decoder, providing a meaningful preview at each generation stage.

\subsection{Interactive Generation and Editing}
\label{sec:editing}

Because generation is decomposed into discrete, interpretable levels, users can inspect, modify, and re-generate at any point along the trajectory.

\noindent \textbf{Editing workflow.}
Given a fully generated trajectory $\{\hat{\mathbf{z}}^{(l)}\}_{l=0}^{L-1}$, a user inspects the decoded output at level $l^*$ and produces an edited canvas $\tilde{\mathbf{z}}^{(l^*)}$.
Generation then resumes from this edit: levels $l > l^*$ are re-generated conditioned on the modified output, while all coarser levels $l < l^*$ remain unchanged.
This workflow supports two complementary editing operations:

\noindent \textbf{(i) Feature editing:}
The user replaces the mean feature of a selected region in $\hat{\mathbf{z}}^{(l^*)}$ with a feature sourced from another region or image (e.g., swapping one part's semantics with another), then propagates forward through subsequent levels.
Because semantically similar content maps to nearby features in DINOv2 space, this transfers the semantic identity of the source region to the target.

\noindent \textbf{(ii) Shape editing:}
The user modifies the spatial extent of a region by reassigning tokens at the boundary between adjacent regions in $\hat{\mathbf{z}}^{(l^*)}$, then propagates forward.
This changes \emph{where} a region is without altering its feature content, for example, enlarging or shrinking a part, or reshaping its contour.

A key property underlying both operations is editing \emph{scope control}. 
Since generation is Markov (each level conditioned only on the preceding one), an edit at level $l^*$ propagates through levels $l > l^*$ but leaves coarser levels untouched.
This gives users predictable control over editing scope: coarser edits cascade through more stages and have broader semantic impact, while finer edits remain spatially localized.
These editing operations motivate the trajectory-aware evaluation metrics introduced in Sec.~\ref{sec:exp}.

%% file: sec/exp.tex
\begin{figure*}[t]
    \centering
    \captionsetup{font={stretch=0.8,footnotesize}}
    \includegraphics[width=0.98\linewidth,keepaspectratio]{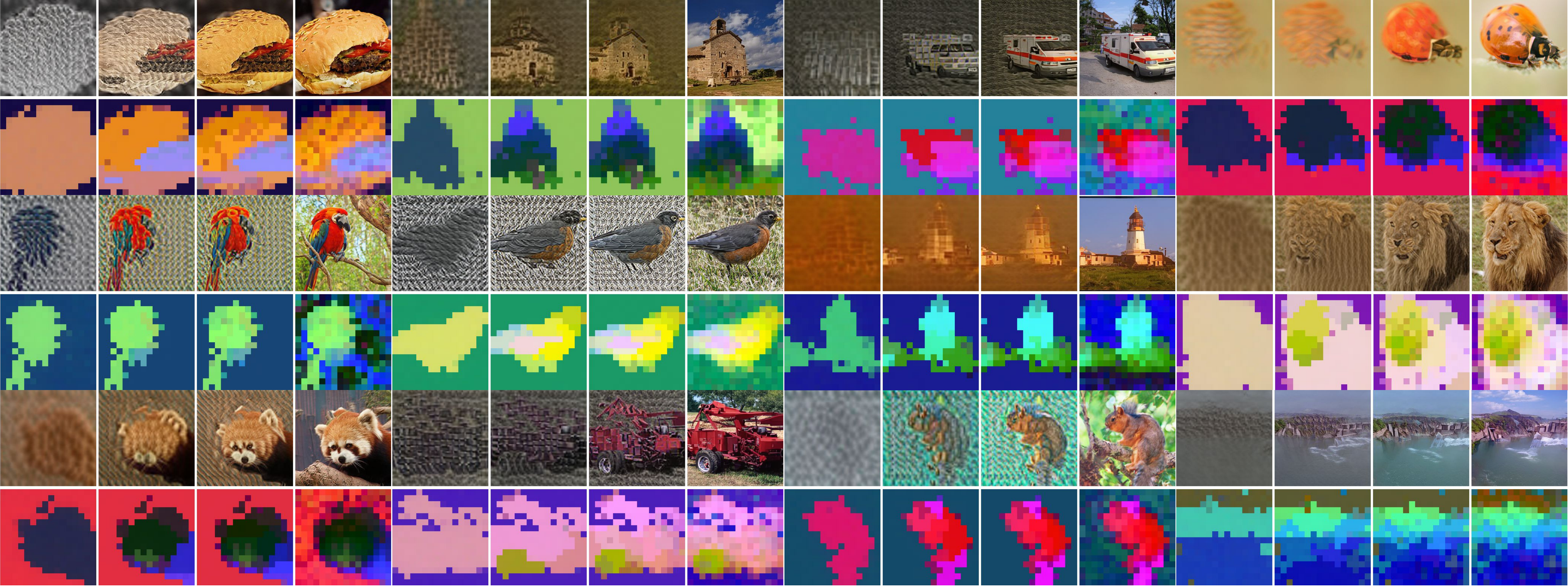}
    \vspace{-2mm}
    \caption{\textbf{Decoded samples} with PCA colored latent stages.}
    \label{fig:samples}
\vspace{-5mm}
\end{figure*}

\section{ImageNet Experiments}
\label{sec:exp}

We evaluate on ImageNet~\cite{imagenet} at $256{\times}256$ resolution.
Following the hierarchical sampling of Sec.~\ref{sec:sampling}, each image is produced in $L{=}4$ one-step denoising stages in DINOv2~\cite{oquab2023dinov2} feature space, with every intermediate output decodable via a pre-trained ViT-XL decoder~\cite{zheng2026rae}.
We report FID~\cite{heusel2017fid} and IS~\cite{salimans2016improved} on 50k samples.

Our backbone is a DiT~\cite{peebles2023scalable} transformer shared across all levels, with patch size $16{\times}16$ (denoted TF/16).
By default, we train TF with structural loss weight $\lambda{=}1$ and the Muon optimizer~\cite{muon} at a constant learning rate of $1e-2$. 
Implementation details, full hyperparameters, scalability across model sizes, and $\lambda$ ablation are provided in the supplementary material.

\input{tables/fid}

\subsection{System-level Comparison}
\label{sec:fid}

We compare with prior methods in Tab.~\ref{tab:model-comparison}. Since work on hierarchical latent generation is scarce, we include representative pixel-space as well as latent-space baselines, spanning both multi-step and one-step diffusion/flow approaches. Unless otherwise noted, we report results for models trained from scratch.

At $80$ training epochs, TF shows rapid convergence relative to baselines that typically require several hundred epochs. We note that these are early-training results; the primary contribution of TF is not FID optimization but trajectory-level controllability, which no other method in the table provides.

\noindent \textbf{Relation to prior work.}
The most related prior work is NVG~\cite{wang2026nvg}, which also performs hierarchical generation; we discuss conceptual differences in Sec.~\ref{sec:related}.
From a practical standpoint, two differences are worth highlighting:
\textbf{(i)}~NVG requires training a custom multi-granularity VQ-VAE 
to support its resolution-based hierarchy, whereas TF operates directly in pretrained DINOv2 features without a task-specific tokenizer;
\textbf{(ii)}~NVG's structure generation relies on a multi-step rectified flow model at each stage ($9$ content steps + $7\times 25$ Euler steps = $184$ NFEs); moreover, its discrete nature makes it difficult to leverage recent one-step advances~\cite{geng2025imf,lu2026pmf,deng2026drifting}, whereas TF naturally integrates one-step matching and produces each level in a single evaluation ($L{=}4$ NFEs total).


\subsection{Evaluation beyond FID}

\label{sec:metrics}

Standard metrics such as FID evaluate distributional quality, but do not capture the controllability and structural faithfulness central to TF.
To measure these properties, we introduce trajectory-aware metrics along two complementary axes: (i)~spatial locality of edits and (ii)~structural coherence across levels.

\begingroup
\raggedbottom
\begin{figure*}[t]
    \centering
    \captionsetup{font={stretch=0.8,footnotesize}}
    \begin{subfigure}[t]{0.49\linewidth}
        \centering
        \includegraphics[width=\linewidth]{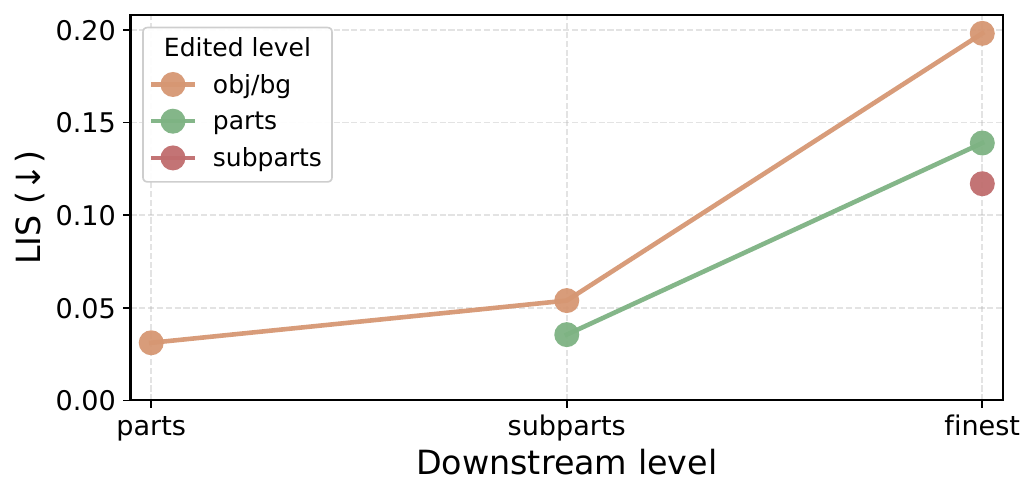}
        \caption{\textbf{Local Invariance Score (LIS)}}
        \label{fig:lis}
    \end{subfigure}\hfill
    \begin{subfigure}[t]{0.49\linewidth}
        \centering
        \includegraphics[width=\linewidth]{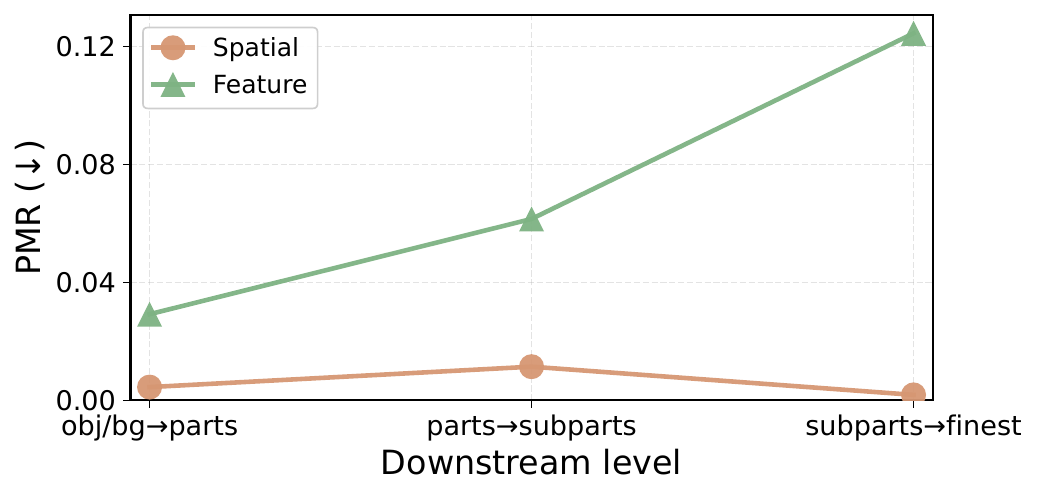}
        \caption{\textbf{Structural Consistency (PMR)}}
        \label{fig:pmr}
    \end{subfigure}
    \caption{\textbf{Trajectory-aware evaluation.}
    (a) LIS shows coarser edits propagate more strongly, while finer edits remain localized.
    (b) Both spatial and feature PMR remain low, indicating hierarchical consistency across levels. 
    We evaluate on TF alone as no baseline produces semantic-region intermediates (NVG uses binary splits over discrete tokens without region structure).
    }
    \label{fig:metrics}
\vspace{-5mm}
\end{figure*}

\subsubsection{5.2.1 Local Invariance under Manipulation.} 

We evaluate whether editing at level $\ell^*$ introduces unintended changes to regions that are not explicitly modified.
Let $\hat{\mathbf{z}}^{(\ell)}$ denote the generated latent at level $\ell$ before manipulation and $\hat{\mathbf{z}}'^{(\ell)}$ the corresponding latent after re-generation from the edited canvas.
Given a spatial mask $\mathbf{M} \in \{0,1\}^{H \times W}$ indicating edited tokens ($M_{ij}{=}1$), we define the \textbf{Latent Invariance Score (LIS)} as the average cosine distance over unedited tokens:
\begin{equation}
    f^{(\ell)}_{\text{lis}}
    =
    \frac{1}{|\Omega_{\text{unedit}}|}
    \sum_{(i,j)\in\Omega_{\text{unedit}}}
    \left( 1 - \cos\!\left<\hat{\mathbf{z}}^{(\ell)}_{i,j},\, \hat{\mathbf{z}}'^{(\ell)}_{i,j}\right> \right),
\end{equation}
where $\Omega_{\text{unedit}} = \{(i,j) \mid M_{ij}{=}0\}$. 
Lower scores indicate stronger local invariance.
We evaluate LIS at each downstream level $\ell \geq \ell^*$ to study how edits propagate through subsequent generation stages.
Fig.~\ref{fig:metrics}(a) shows a clear scale-dependent behavior: coarser edits produce larger downstream deviations, whereas finer edits remain more localized.
Object/background edits yield the highest LIS at the final level, followed by parts and subparts edits.
This confirms that TF exposes a controllable hierarchy where the spatial extent of changes can be modulated by the editing level.

\subsubsection{5.2.2 Structural Consistency across Levels.}

Beyond editing invariance, we evaluate structural coherence between consecutive levels: each child-level region should be spatially contained within a single parent region; a child straddling multiple parents signals structural drift.

To formalize this, given generated outputs at consecutive levels $\ell$ and $\ell{+}1$, we cluster each into regions using the procedure of Sec.~\ref{sec:teacher}.
At level $\ell$, we obtain $K_\ell$ parent regions with mean-feature centers $\{\boldsymbol{\mu}^{(\ell)}_i\}_{i=1}^{K_\ell}$ and per-token assignments $a^{(\ell)}(n) \in \{1,\dots,K_\ell\}$.
At level $\ell{+}1$, we obtain $K_{\ell+1}$ child regions $\{C_j\}_{j=1}^{K_{\ell+1}}$.

Since a child region may partially overlap multiple parent regions near boundaries, we assign each child a parent by spatial majority:
\[
    \mathrm{parent}(j) = \arg\max_{i}\, \left|\{n \in C_j : a^{(\ell)}(n) = i\}\right|.
\]
This induces a token-level expected parent: every token $n$ in child region $j$ inherits $\mathrm{parent}(j)$ as its expected parent.
We define two complementary token-level \textbf{Parent Misrouting Rates (PMR)}:

\noindent \textbf{(i) Spatial PMR.}
The fraction of tokens whose parent-level cluster assignment disagrees with the majority-voted parent of their child region:
\begin{equation}
    f_\text{pmr-s}^{(\ell)}
    =
    \mathbb{P}_{n}
    \left[
    a^{(\ell)}(n) \neq \mathrm{parent}(j_n)
    \right],
\end{equation}
where $j_n$ denotes the child region containing token $n$.
This measures whether child regions are cleanly contained within single parent regions.

\noindent \textbf{(ii) Feature PMR.}
The fraction of tokens whose feature is closer to an incorrect parent center than to the assigned one:
\begin{equation}
    f_\text{pmr-f}^{(\ell)}
    =
    \mathbb{P}_{n}
    \left[
    \arg\max_{i}\,
    \cos\!\left<
    \hat{\mathbf{z}}^{(\ell+1)}_{n},\,
    \boldsymbol{\mu}^{(\ell)}_i
    \right>
    \neq \mathrm{parent}(j_n)
    \right].
\end{equation}
This captures whether generation has shifted token features away from their assigned parent's semantic mode. 

Lower values indicate stronger structural consistency; both metrics are evaluated at each consecutive level pair.
Fig.~\ref{fig:metrics}(b) shows that spatial PMR remains near zero across levels, indicating that child regions are largely contained within their parent regions.
Feature PMR is slightly higher and increases at deeper levels due to finer semantic partitioning, but remains low overall.
These results indicate that TF preserves hierarchical consistency while progressively refining semantic detail.
\endgroup

\noindent \textbf{Edit effectiveness and decoded-space checks.}
The metrics above verify that edits do not disrupt unedited regions and that generation is structurally coherent.
We additionally evaluate whether edits achieve their intended effect in the target region, and verify that latent-level properties translate to pixel space by decoding intermediate outputs via the frozen RAE decoder.
Editing examples and decoded-space analysis are provided in the supplementary material.

\noindent \textbf{Why a shared decoder?}
A natural question is whether a single decoder suffices for all levels, since intermediate canvases differ in distribution from the final latent.
We deliberately avoid per-level finetuning for two reasons: first, a shared decoder ensures that decoded images across levels live in a consistent visual space, which is essential for editing: a user must be able to compare the output at level $\ell^*$ with the final image; second, no ground-truth intermediate images exist, as level canvases are piecewise-constant abstractions in feature space, making per-level supervision unavailable by construction.

%% file: tables/fid.tex
\begin{table*}[t]

    \centering
    \scriptsize
    \setlength{\tabcolsep}{2pt}
    \renewcommand{\arraystretch}{1.0}

    \caption{
        \textbf{Comparison on ImageNet 256$\times$256.}
        FID and IS are computed on 50k samples with CFG where applicable (${\times}2$ on NFE indicates CFG doubles the cost).
        We compare against baselines of comparable model size and, when applicable, similar training epochs; extended comparisons can be found in in the supplementary.
    }

    \label{tab:model-comparison}
    \vspace{-3mm}
    \begin{tabularx}{\linewidth}{X >{\scriptsize}c c >{\scriptsize}c >{\scriptsize}c >{\scriptsize}c >{\scriptsize}c}

        \noalign{\hrule height 1pt}
        \noalign{\vskip 0.1em}

        \textbf{ImgNet 256$\times$256} 
        & \textbf{NFE} 
        & \textbf{Space} 
        & \textbf{Params} 
        & \textbf{Epochs}
        & \textbf{FID($\downarrow$)} 
        & \textbf{IS($\uparrow$)} \\

        \noalign{\hrule height 1pt}

        \rowcolor{black!10}\multicolumn{7}{l}{\textbf{\textit{Multi-step pixel diffusion/flow}}} \\

        JiT-B/16 (Heun)~\cite{li2026jit}  
        & $100\times2$ & pixel & 131M & 200 & 4.37 & 275.1 \\

        JiT-L/16 (Heun)  
        & $100\times2$ & pixel & 459M & 200 & 2.36 & 298.5 \\

        DeCo-XL/16~\cite{ma2025deco} 
        & $100\times2$ & pixel & 682M & 320 & 1.90 & 303.0 \\

        LF-ViT/L (Heun)~\cite{baade2026latentforcing} 
        & $50\times2$ & pixel+DINOv2 & 465M & 200 & 2.48 & -- \\

        \noalign{\hrule height 1pt}

        \rowcolor{black!10}\multicolumn{7}{l}{\textbf{\textit{Multi-step latent diffusion/flow}}} \\

        DiT-B/2~\cite{peebles2023scalable}             
        & 250 & SD-VAE & 130M & 80 & 43.47 & 278.2 \\

        DiT-XL/2             
        & 250 & SD-VAE & 675M & 1400 & 9.62 & 121.5 \\

        DiT-XL/2 (cfg=1.50)             
        & $250\times2$ & SD-VAE & 675M & 1400 & 2.27 & 278.2 \\

        SiT-B/2~\cite{ma2024sit}+REPA~\cite{yu2025repa}     
        & 250 & SD-VAE & 130M & 80 & 24.40 & -- \\

        SiT-XL/2             
        & 250 & SD-VAE & 675M & 1400 & 8.61 & 131.7 \\

        SiT-XL/2 (cfg=1.50)             
        & $250\times2$ & SD-VAE & 675M & 1400 & 2.06 & 270.3 \\

        RAE+DiT$^{\mathrm{DH}}$-XL/2~\cite{zheng2026rae} 
        & $50\times2$  & DINOv2 & 839M & 800 & 1.13 & 262.6 \\

        \noalign{\hrule height 1pt}

        \rowcolor{black!10}\multicolumn{7}{l}{\textbf{\textit{Autoregressive latent diffusion/flow}}} \\

        VAR-d16~\cite{tian2024var}   
        & 250 & VQ-VAE & 310M & 200 & 3.30 & 274.4 \\

        VAR-d20 
        & 250 & VQ-VAE & 600M & 250 & 2.57 & 302.64 \\

        NVG-d16~\cite{wang2026nvg}   
        & 184\textsuperscript{*} & VQ-VAE & 255M & 200 & 3.03 & 279.2 \\

        NVG-d20 
        & 184\textsuperscript{*} & VQ-VAE & 497M & 250 & 2.44 & 310.4 \\

        \noalign{\hrule height 1pt}
        \rowcolor{black!5}\multicolumn{7}{l}{\textbf{\textit{1-NFE diffusion/flow}}} \\

        pMF-B/16~\cite{lu2026pmf} 
        & 1 & pixel & 119M  & 320 & 3.12 & 254.6 \\

        pMF-L/16 
        & 1 & pixel & 411M & 320 & 2.52 & 262.6 \\

        EPG-B/16~\cite{lei2026novae}            
        & 1 & pixel & 229M & 240 & 25.10 & -- \\

        EPG-L/16~\cite{lei2026novae}            
        & 1 & pixel & 540M & 560 & 8.82 & -- \\

        \hline

        iCT-XL/2~\cite{song2024improved}            
        & 1 & SD-VAE  & 675M & 240 & 34.24\textsuperscript{\textdagger} & -- \\

        Shortcut-XL/2~\cite{frans2025one}       
        & 1 & SD-VAE  & 676M & 240 & 10.60\textsuperscript{\textdagger} & 102.7 \\

        IMM-XL/2~\cite{zhou2025imm}       
        & 1$\times$2 & SD-VAE  & 676M & 240 & 7.77\textsuperscript{\textdagger} & -- \\

        MeanFlow-B/4~\cite{geng2025mf}       
        & 1 & SD-VAE & 131M  & 80 & 15.53  & -- \\

        iMF-M/2~\cite{geng2025imf}            
        & 1 & SD-VAE & 174M  & 640 & 2.27  & -- \\

        iMF-L/2~\cite{geng2025imf}            
        & 1 & SD-VAE & 409M  & 640 & 1.86  & -- \\

        \hline

        \rowcolor{black!5}\textbf{TF-B/16 (ours)} 
        & 4$\times$1 & DINOv2 & 177M & 80 & 7.93 & 221.4 \\

        \rowcolor{black!5}\textbf{TF-B/16$\mathtt{+}$FD-loss\textsuperscript{\textdaggerdbl} (ours)} 
        & 4$\times$1 & DINOv2 & 515M & 80 & 2.52 & 245.6 \\
        
        \rowcolor{black!5}\textbf{TF-L/16 (ours)} 
        & 4$\times$1 & DINOv2 & 177M & 80 & 6.38 & 248.1 \\

        \rowcolor{black!5}\textbf{TF-L/16$\mathtt{+}$FD-loss\textsuperscript{\textdaggerdbl} (ours)} 
        & 4$\times$1 & DINOv2 & 515M & 80 & 2.02 & 259.3 \\

        \rowcolor{black!5}\textbf{TF-H/16 (ours)} 
        & 4$\times$1 & DINOv2 & 1.1B & 80 & 5.97 & 256.8 \\

        \rowcolor{black!5}\textbf{TF-H/16$\mathtt{+}$FD-loss\textsuperscript{\textdaggerdbl} (ours)} 
        & 4$\times$1 & DINOv2 & 1.1B & 80 & 1.98 & 261.6 \\

        \noalign{\hrule height 1pt}
        \noalign{\vskip 0.3em}
        \multicolumn{7}{l}{\tiny \textsuperscript{*}NVG requires 184 NFEs total; see text Sec.~\ref{sec:fid} for breakdown.} \\[0.1em]
        \multicolumn{7}{l}{\tiny \textsuperscript{\textdagger}Results reported from the original MeanFlow paper (trained from scratch).} \\[0.1em]
        \multicolumn{7}{l}{\tiny \textsuperscript{\textdaggerdbl}Inception-space FD-loss~\cite{yang2026representation} post-training; see text App.~\ref{sec:implementation_details} for details.} 
    
    \end{tabularx}
    \vspace{-5mm}

\end{table*}

%% file: appendix.tex
\begin{center}
    {\Large\bfseries Appendix\par}
\end{center}
\vspace{1em}

\appendix

\makeatletter
\renewcommand*{\theHsection}{appendix.\Alph{section}}
\renewcommand*{\theHsubsection}{appendix.\Alph{section}.\arabic{subsection}}
\renewcommand*{\theHsubsubsection}{appendix.\Alph{section}.\arabic{subsection}.\arabic{subsubsection}}
\makeatother

\setcounter{tocdepth}{3}
\makeatletter
\begingroup
\noindent\textbf{\large Table of Contents}\par
\vspace{0.3em}
\hrule
\vspace{0.3em}
\let\oldl@section\l@section
\renewcommand*{\l@section}[2]{\oldl@section{\bfseries #1}{#2}}
\@starttoc{aptoc}
\vspace{0.3em}
\hrule
\endgroup
\let\tf@addcontentsline\addcontentsline
\renewcommand{\addcontentsline}[3]{%
  \tf@addcontentsline{#1}{#2}{#3}%
  \def\tf@tocname{#1}%
  \def\tf@targettoc{toc}%
  \ifx\tf@tocname\tf@targettoc
    \addtocontents{aptoc}{\protect\contentsline{#2}{#3}{\thepage}{\@currentHref}}%
  \fi
}
\makeatother

\newpage

\section{Experimental Settings}

\subsection{Additional Implementation Details}
\label{sec:implementation_details}

\noindent \textbf{Architecture.}
Following pMF~\cite{lu2026pmf}, our network uses a shared backbone with split u/v prediction heads.
The previous-level canvas is fused via channel-wise concatenation: both the noisy input and the conditioning canvas are independently projected to the hidden dimension, concatenated along the channel axis, and projected back via a linear layer.
The level index is encoded through a learned embedding table (4 tokens) and appended to the conditioning token sequence alongside class and timestep. 

\noindent \textbf{Data preprocessing.}
Images are center-cropped to $256{\times}256$, normalized to $[-1, 1]$, and augmented with random horizontal flips (50\%).
DINOv2 features and hierarchical region maps (object/background, parts, subparts) are precomputed offline and stored alongside images.

\noindent \textbf{Decoder.}
We use a frozen ViT-XL RAE decoder~\cite{zheng2026rae} pretrained on DINOv2 features.
The decoder maps $16{\times}16$ latent grids with 768-dimensional tokens back to $256{\times}256$ RGB images.
No per-level decoder finetuning is performed; the same decoder is used for all hierarchy levels.

\noindent \textbf{EMA.}
We maintain three exponential moving averages with EDM-style scheduling~\cite{karras2024edm2} and select the best-performing one during inference.

\noindent \textbf{Classifier-free guidance.} We disable classifier-free guidance (CFG) throughout training and inference. Unless otherwise stated, all results are reported with a guidance scale of $\omega = 1.0$.

\noindent \textbf{Post-training with Fréchet Distance (FD) loss.}
Since FID remains a widely reported benchmark metric, TF can optionally adopt the Inception-space FD-loss post-training procedure~\cite{yang2026representation} to improve terminal-image fidelity under the conventional FID protocol. 
This stage is orthogonal to our trajectory objective and does not change the proposed trajectory-structured training.

We also stress that Inception-space FD/FID should be interpreted only as a proxy for perceptual quality. Prior work has shown that FID can be sensitive to the choice of feature representation, finite-sample estimation, and evaluation protocol, and may disagree with human perceptual judgment~\cite{kynkaanniemi2023role,jayasumana2024rethinking,chong2020effectively,parmar2022aliased}. Consistently,~\cite{yang2026representation} observes that optimizing FD in modern representation spaces can yield visually better samples despite worse Inception FID. Therefore, we use FD-loss post-training only as an optional metric-oriented refinement of final images, while our main contribution remains structuring and controlling the intermediate generative trajectory.

\subsection{Hyperparameter Settings}
The full configurations are given in \Cref{tab:config-hparams}. Our implementation is based on pMF~\cite{lu2026pmf}, using JAX on 8$\times$ H200 GPUs.
We highlight several key choices below.

\noindent \textbf{Timestep sampling.} Following Self-Flow~\cite{chefer2026self}, we sample the timesteps $(t,r)$ using a plateau-logit-normal distribution parameterized by a mean of $0.0$, a standard deviation of $1.0$, and a shift factor of $\alpha = 10.0$.

\noindent \textbf{Structural loss.}
We set $\lambda{=}1$ for the structural loss weight across all model sizes. The structural loss is applied only to levels $l \in \{0, \ldots, L{-}2\}$ and disabled at the finest level.

\noindent \textbf{Model scaling.} We train three model sizes: TF-B/16 (177M parameters), TF-L/16 (515M), and TF-H/16 (1.1B). We train TF-B/16 in fp32 for the ablation studies for 80 epochs, whereas TF-L/16 and TF-H/16 are trained in bf16 for efficiency. TF-B/16 and TF-L/16 use a global batch size of 1024, while TF-H/16 uses a global batch size of 512.

\input{tables/config}

\section{Editing Analysis}
\label{sec:editing_supp}

A central claim of Trajectory Forcing is that the structured generative trajectory enables interactive control: users can inspect intermediate states, apply targeted modifications, and resume generation from the edit point.
In the main paper, we describe two editing operations (feature editing and shape editing) and quantify their locality via LIS and structural coherence via PMR.
Here we provide step-by-step procedures, qualitative examples at multiple hierarchy levels, and decoded-space sanity checks.

\subsection{Editing Workflow}
\label{sec:supp_workflow}

Since generation proceeds sequentially from coarse to fine, editing is naturally interleaved with generation.
At each level $l$, the user performs three steps:

\begin{enumerate}
    \item \textbf{Generate.} Produce $\hat{\mathbf{z}}^{(l)}$ from noise conditioned on the (possibly edited) output of the previous level (or from zero conditioning at $l{=}0$).
    \item \textbf{Inspect.} Decode $\hat{\mathbf{z}}^{(l)}$ via the shared RAE decoder to obtain a visual \emph{preview} at the current granularity. This preview reveals the semantic content decided so far, allowing the user to assess whether the layout, part assignment, or detail matches their intent before committing to finer levels.
    \item \textbf{Edit (optional).} If the user wishes to intervene, modify $\hat{\mathbf{z}}^{(l)}$ to produce an edited canvas $\tilde{\mathbf{z}}^{(l)}$. The specific modification depends on the editing type (feature replacement or shape change; see below). All unedited tokens remain unchanged.
\end{enumerate}

\noindent The output of each level (edited or not) serves as the conditioning input for the next, so modifications are automatically propagated through all subsequent levels.
This workflow is fully deterministic given a fixed noise seed, making edits reproducible and comparable.

\subsection{Feature Editing}

\begin{figure*}[t]
    \centering
    \includegraphics[width=1.0\linewidth]{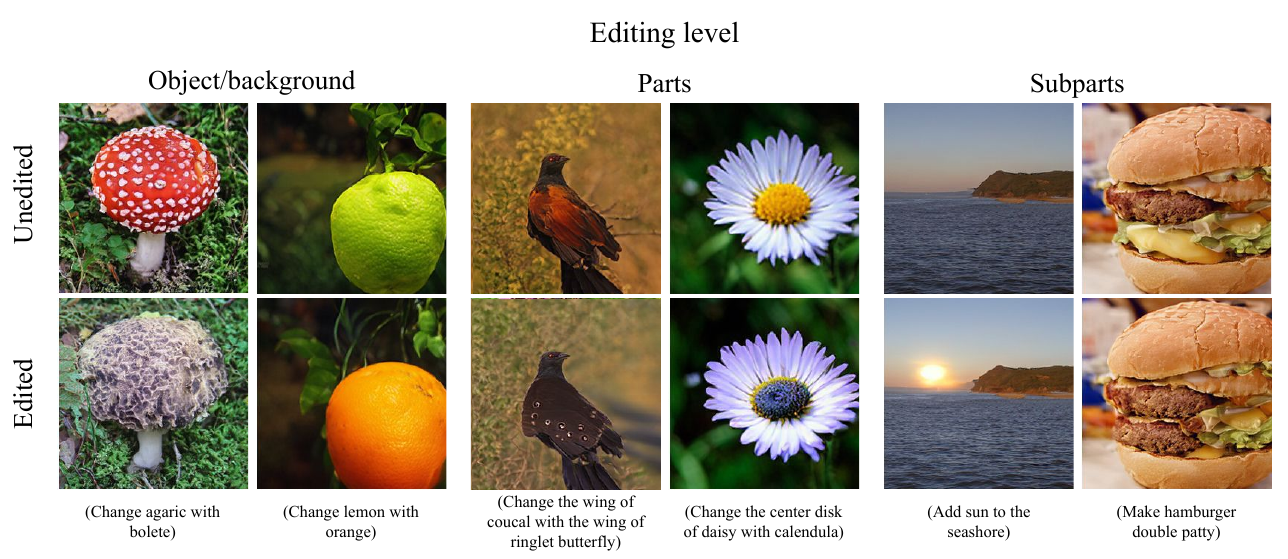}
    \vspace{-2mm}
    \captionsetup{font={stretch=0.9,footnotesize}}
    \captionof{figure}{ \textbf{Feature editing at different hierarchy levels.} The top row shows the original generated images, and the bottom row shows results after feature editing. A target region is selected and its latent tokens are replaced with a source feature token. Editing at the object/background level changes the global semantic identity of the scene while preserving spatial layout. Editing at the part level modifies a single object part while keeping the rest of the object unchanged. Editing at the subpart level produces localized edits affecting only fine-grained details. This demonstrates that editing earlier hierarchy levels produces broader semantic changes, while deeper levels enable precise local control.
    }
    \label{fig:feature_edit}
\end{figure*}

Feature editing changes \emph{what} a region represents while preserving its spatial layout.

\noindent \textbf{Procedure.}
Given the generated canvas $\hat{\mathbf{z}}^{(l^*)}$, the user selects a target region $R_{\text{tgt}}$ and a source feature $\mathbf{f}_{\text{src}}$.
The source can be: (a)~the mean feature of another region in the same image (e.g., swapping the semantics of two parts), or (b)~a mean feature extracted from a different image (e.g., transferring the appearance of a specific object part from a reference).
The edit replaces all tokens in $R_{\text{tgt}}$ with $\mathbf{f}_{\text{src}}$:
\[
    \tilde{\mathbf{z}}^{(l^*)}_i =
    \begin{cases}
        \mathbf{f}_{\text{src}} & \text{if } i \in R_{\text{tgt}}, \\
        \hat{\mathbf{z}}^{(l^*)}_i & \text{otherwise}.
    \end{cases}
\]
Because DINOv2 features are semantically organized, nearby features correspond to similar visual content.
Replacing a region's feature therefore transfers the semantic identity of the source to the target location, while the spatial structure (region shape and surrounding context) is preserved.

\noindent \textbf{Effect at different levels.}
Editing at the object/background level ($l^*{=}0$) swaps the global semantic identity of the foreground or background, causing all downstream levels to regenerate with the new identity: the broadest possible semantic change.
Editing at the part level ($l^*{=}1$) replaces a single part (e.g., changing the wing of a bird) while keeping other parts and the overall object layout intact.
Editing at the subpart level ($l^*{=}2$) produces the most localized change, affecting only fine-grained details within a subregion.

\subsection{Shape Editing}

\begin{figure*}[t]
    \centering
    \includegraphics[width=1.0\linewidth]{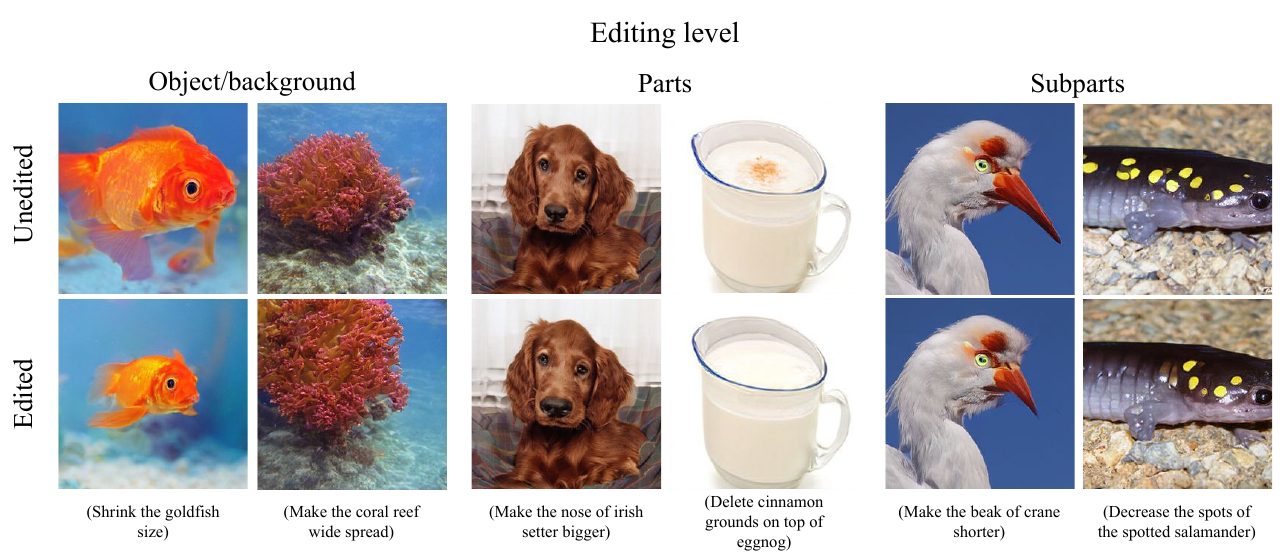}
    \vspace{-2mm}
    \captionsetup{font={stretch=0.9,footnotesize}}
    \captionof{figure}{ \textbf{Shape editing at different hierarchy levels.} The top row shows the original generated images, and the bottom row shows results after shape editing. A boundary between two adjacent regions is modified by reassigning tokens from one region to the other. Editing at the object/background level changes the coarse spatial allocation between foreground and background. Editing at the part level reshapes individual object parts while preserving their semantic identity. Editing at the subpart level produces localized contour adjustments. Because only region assignments are changed, semantic feature content remains unchanged while spatial structure is modified.}
    \label{fig:shape_edit}
\end{figure*}

Shape editing changes \emph{where} a region is without altering its semantic content.

\noindent \textbf{Procedure.}
The user selects a boundary between two adjacent regions $R_a$ and $R_b$ in $\hat{\mathbf{z}}^{(l^*)}$ and reassigns boundary tokens from one region to the other.
Concretely, for each reassigned token $i$ moved from $R_a$ to $R_b$:
\[
    \tilde{\mathbf{z}}^{(l^*)}_i = \boldsymbol{\mu}_{R_b},
\]
where $\boldsymbol{\mu}_{R_b}$ is the mean feature of the receiving region.
This enlarges $R_b$ at the expense of $R_a$ (or vice versa), effectively reshaping region contours.
The feature content of both regions remains unchanged; only the spatial extent is modified.

\noindent \textbf{Effect at different levels.}
At the object/background level, shape editing adjusts the coarse spatial allocation between foreground and background. For instance, enlarging the object region or shifting its position.
At the part level, it reshapes individual parts (e.g., making a nose bigger), which then propagates to finer subpart structure in downstream levels.
At the subpart level, the change is confined to local contour adjustments.

\subsection{Editing Scope Control}

\begin{figure*}[t]
    \centering
    \includegraphics[width=1.0\linewidth]{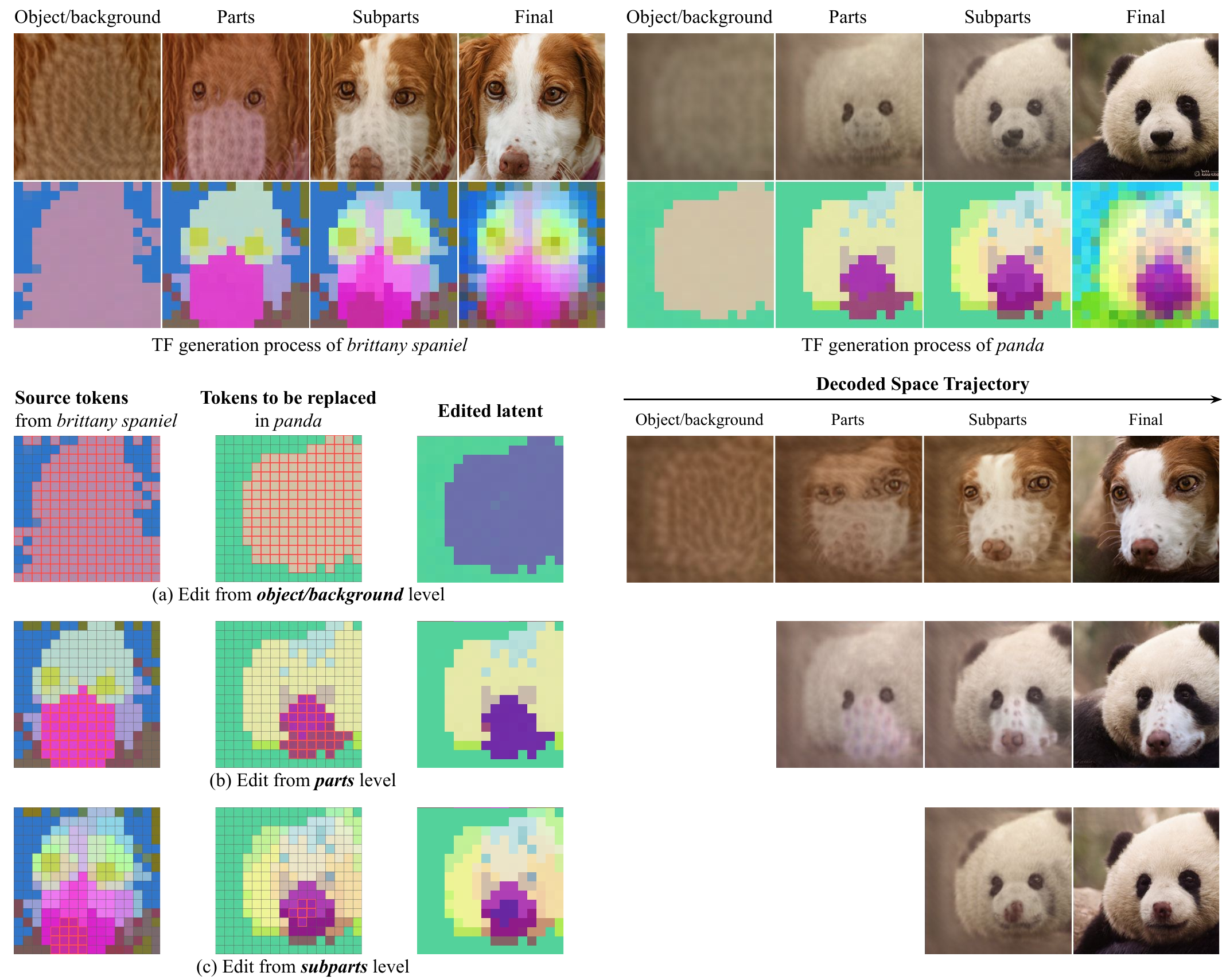}
    \vspace{-4mm}
    \captionsetup{font={stretch=0.9,footnotesize}}
    \captionof{figure}{\textbf{Editing scope control.} The same feature replacement operation is applied at different hierarchy levels on the same generated image. Red boxes indicate the selected region. Editing at \textbf{(a) the object/background level} produces a global semantic change; editing at \textbf{(b) the part level} modifies a single part while preserving overall layout; editing at \textbf{(c) the subpart level} results in a spatially localized change. This demonstrates that the editing level directly controls the breadth of downstream impact.}
    \label{fig:editing_scope_control}
\end{figure*}

A distinctive property of TF editing is predictable \emph{scope control}: the choice of editing level $l^*$ directly determines the breadth of downstream impact.
Since generation is Markov (each level conditioned only on the immediately preceding one), an edit at $l^*$ propagates through all levels $l > l^*$ but leaves all coarser levels $l < l^*$ untouched.

To illustrate this, we apply the same type of edit (\emph{feature replacement} of a single region) at three different levels: object/background ($l^*{=}0$), part ($l^*{=}1$), and subpart ($l^*{=}2$), and compare the resulting changes in the final decoded image.
As shown in Fig.~\ref{fig:editing_scope_control}, editing at $l^*{=}0$ causes global semantic change across the entire image; editing at $l^*{=}1$ alters one part while preserving overall layout; editing at $l^*{=}2$ produces a spatially localized change confined to a small subregion.
This directly corresponds to the quantitative LIS results in the main paper Fig.~6(a), where coarser edits produce progressively larger downstream deviations.

In practice, we provide an interactive interface where users can either select individual tokens or automatically group a contiguous region by cosine similarity to a seed token, making editing straightforward at any granularity.

\subsection{Decoded-space Sanity Checks}

\input{tables/decoded_sanity_checks}

The trajectory-aware metrics in the main paper (LIS, PMR) operate in DINOv2 feature space.
To verify that latent-level invariance translates to pixel space, we decode generated outputs before and after editing using the frozen RAE decoder and compute two standard perceptual metrics restricted to unedited regions:
\begin{itemize}
    \item \textbf{Masked SSIM}: structural similarity computed over the full image, then averaged only over unedited pixels $\Omega_{\text{unedit}}$.
    \item \textbf{Masked LPIPS}: perceptual distance computed over the full image, then averaged over $\Omega_{\text{unedit}}$.
\end{itemize}
As shown in \Cref{tab:decoded_sanity}, high masked SSIM and low masked LPIPS confirm that unedited regions remain visually unchanged after editing, validating that the latent-level locality measured by LIS is preserved through decoding.

\section{Additional Ablation Studies}
We ablate additional important factors below.

\subsection{Structural Loss Weight}

The structural loss $\mathcal{L}_\text{struct}$ encourages region-level consistency, which is also a prerequisite for reliable hierarchical editing. Without coherent region structure, edits applied at a given level may not propagate correctly to downstream levels. However, an excessively large weight $\lambda$ can interfere with the flow objective $\mathcal{L}_\text{flow}$ and degrade generative quality.

\begin{figure*}
    \vspace{-2mm}
    \centering
    \captionsetup{font={stretch=0.9,footnotesize}}
    
    \begin{subfigure}[t]{0.49\linewidth}
        \centering
        \includegraphics[width=\linewidth]{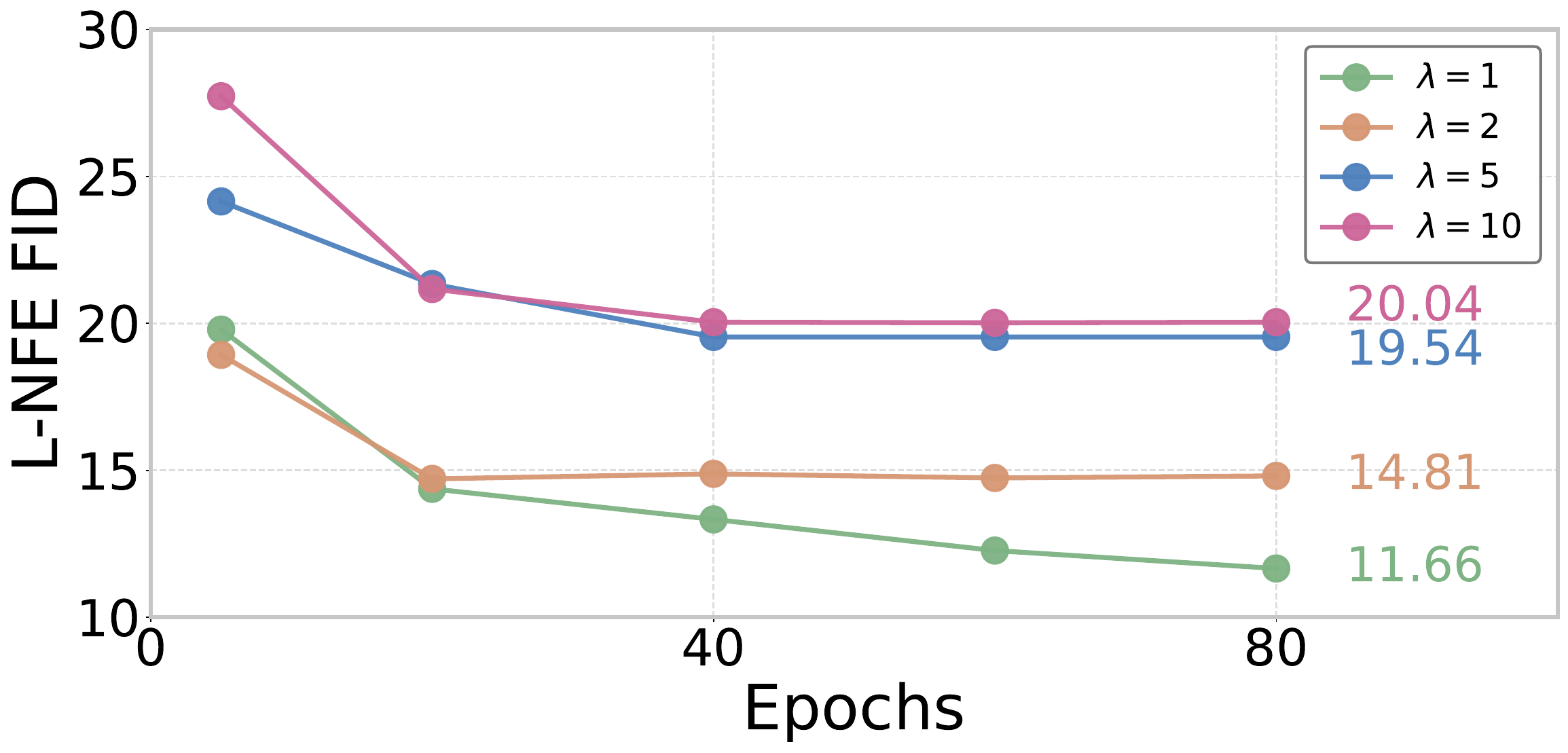}
        \caption{\textbf{Structural loss weight ablation.}}
        \label{fig:loss_abl}
    \end{subfigure}
    \hfill
    \begin{subfigure}[t]{0.49\linewidth}
        \centering
        \includegraphics[width=\linewidth]{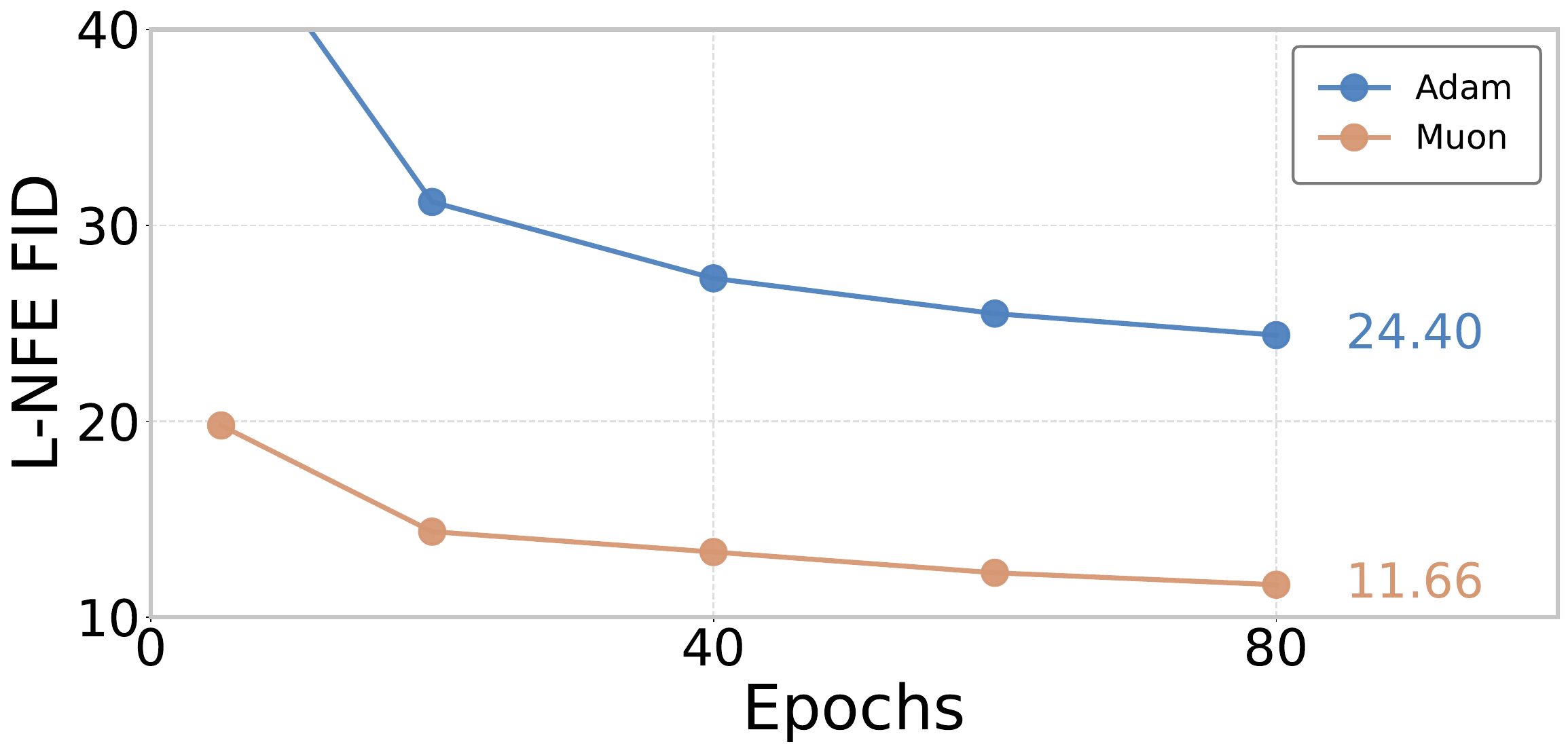}
        \caption{\textbf{Optimizer ablation (AdamW vs. Muon).}}
        \label{fig:opt_abl}
    \end{subfigure}
    \vspace{-2mm}
    
    \captionsetup{font={stretch=0.9,footnotesize}}
    \caption{\textbf{Ablation studies} on TF-B/16. (a) We sweep $\lambda \in \{1,2,5,10\}$ and report $L$-NFE FID throughout training. Smaller weights lead to better final sample quality: $\lambda{=}1$ achieves the lowest FID, while larger weights converge earlier but plateau at worse performance.
    (b) Optimizer comparison. Under identical training settings, Muon consistently achieves lower $L$-NFE FID than Adam, indicating faster convergence.}
    \label{fig:ablations}
    \vspace{-6mm}
\end{figure*}

We therefore perform an ablation over $\lambda \in \{1,2,5,10\}$ and track the resulting FID during training, as shown in Fig.~\ref{fig:ablations}(a). Smaller weights consistently produce better results: $\lambda=1$ yields the lowest final FID and continues improving throughout training. In contrast, larger weights ($\lambda=5,10$) converge quickly but plateau at significantly worse FID values, indicating that overly strong structural constraints hinder the generative objective.

Based on these results, we use $\lambda=1$ for all experiments in the main paper and supplementary.

\subsection{Optimizer: AdamW vs. Muon} 

Recent work on one-step flow matching~\cite{lu2026pmf} has shown that the Muon optimizer~\cite{muon} yields substantially faster FID convergence than AdamW for flow-based generation. We verify whether this advantage carries over to the hierarchical setting of TF by training TF-B/16 under identical conditions with both optimizers.

As shown in Fig.~\ref{fig:ablations} (b), Muon consistently achieves lower $L$-NFE FID throughout training compared to AdamW, matching observations in prior work on one-step flow matching~\cite{lu2026pmf} and suggesting that Muon is better suited for the trajectory prediction objective used in TF.

Based on these results, we adopt Muon as the default optimizer for all experiments reported in this work.

\section{Extended Discussion}
\label{sec:discussion_supp}

\subsection{Interpretability of Intermediate States}

A core premise of TF is that every intermediate level can be decoded into a visually meaningful image.
This property underpins the inspect step of the editing workflow (Sec.~\ref{sec:supp_workflow}): the user relies on decoded previews to decide whether and where to intervene.
In practice, coarser levels produce piecewise-constant feature maps (region means), which lie outside the typical distribution of DINOv2 encodings.
Despite this distribution shift, the shared RAE decoder produces recognizable outputs at all levels ($L{=}4$), likely because region-mean features still reside in the convex hull of naturally occurring token features.
However, for very coarse levels over highly cluttered scenes, decoded images may lose fine structural cues, reducing their utility for user inspection.

\subsection{Parameter Sharing across Levels}

\noindent \textbf{Observation: level-specific target distributions.}
By construction, the target canvases at different hierarchy levels occupy distinct regions of feature space: coarse levels are piecewise-constant region means, while the finest level consists of original DINOv2 tokens.
These level-specific targets effectively define different target manifolds with different statistical properties.
In the current design, a single shared network learns to map noise to all of these manifolds, distinguished only by the level index and the previous-level conditioning canvas.
While this full sharing is parameter-efficient and already achieves competitive FID, the distributional gap between levels suggests room for improvement through more specialized architectures.

\noindent \textbf{Future directions.}
A natural extension is to relax the degree of parameter sharing.
Concrete strategies include level-specific prediction heads (shared backbone, per-level output projection), lightweight per-level adapters (e.g., LoRA-style modules injected at each transformer block), or fully independent models per level at the cost of increased parameters.
Exploring this sharing-specialization trade-off is a promising direction for improving hierarchical generation.

\subsection{On Generation Beyond Reconstruction}
Modern generative models are 
primarily evaluated 
on photorealistic reconstruction, yet visual creation extends well beyond pixel-level fidelity.
As discussed in the main paper (Fig.~3),  
artists work in terms of color relationships, spatial structure, and perceptual coherence rather than material reproduction -- and recognizable imagery emerges long before fine details are rendered.

This intuition is reflected in the design of TF.
TF's intermediate states are a concrete instance of generation beyond reconstruction: coarse-level outputs capture object layout and part structure without photographic detail, yet remain semantically interpretable and useful for downstream editing.
The fact that users can inspect, evaluate, and redirect generation at these abstract levels suggests that photorealism is not a prerequisite for meaningful visual output -- structure alone carries substantial information.

We believe this perspective opens a broader design space for generative models, where the trajectory itself, not only the final image, serves as a useful creative artifact.
Developing evaluation criteria and interfaces that embrace structural generation beyond endpoint fidelity is an interesting direction for future work.

\clearpage
\section{Additional Qualitative Results}

\begin{figure*}[ht]
    \vspace{-2mm}
    \centering
    \includegraphics[width=\linewidth]{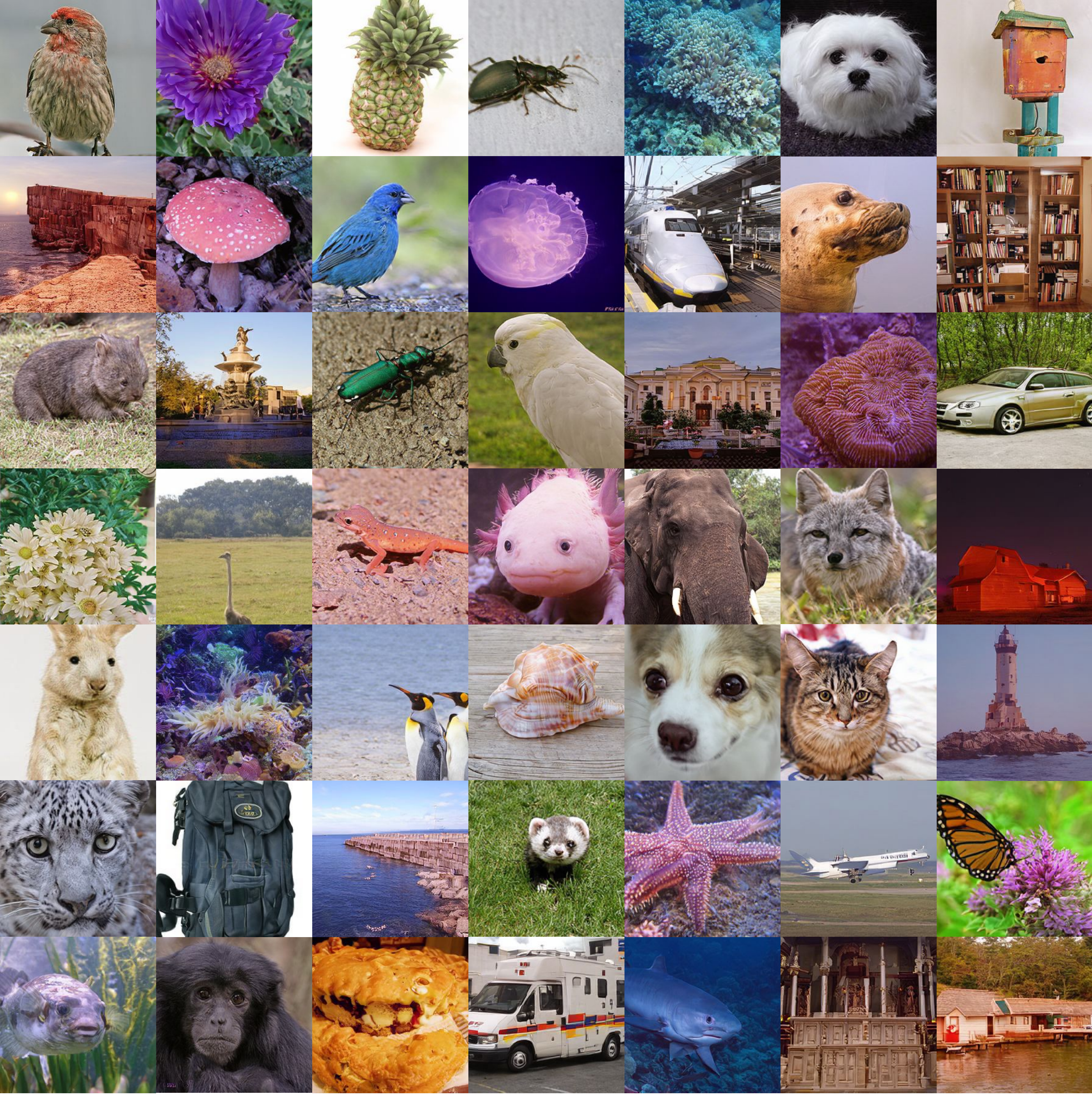}
    \vspace{-2mm}
    \captionsetup{font={stretch=0.9,footnotesize}}
    \captionof{figure}{Uncurated 4-NFE class-conditional generation samples of TF-L/16 on ImageNet 256$\times$256.}
    \label{fig:images_only}
    \vspace{-6mm}
\end{figure*}

\begin{figure*}[ht]
    \vspace{-2mm}
    \centering
    \includegraphics[width=\linewidth]{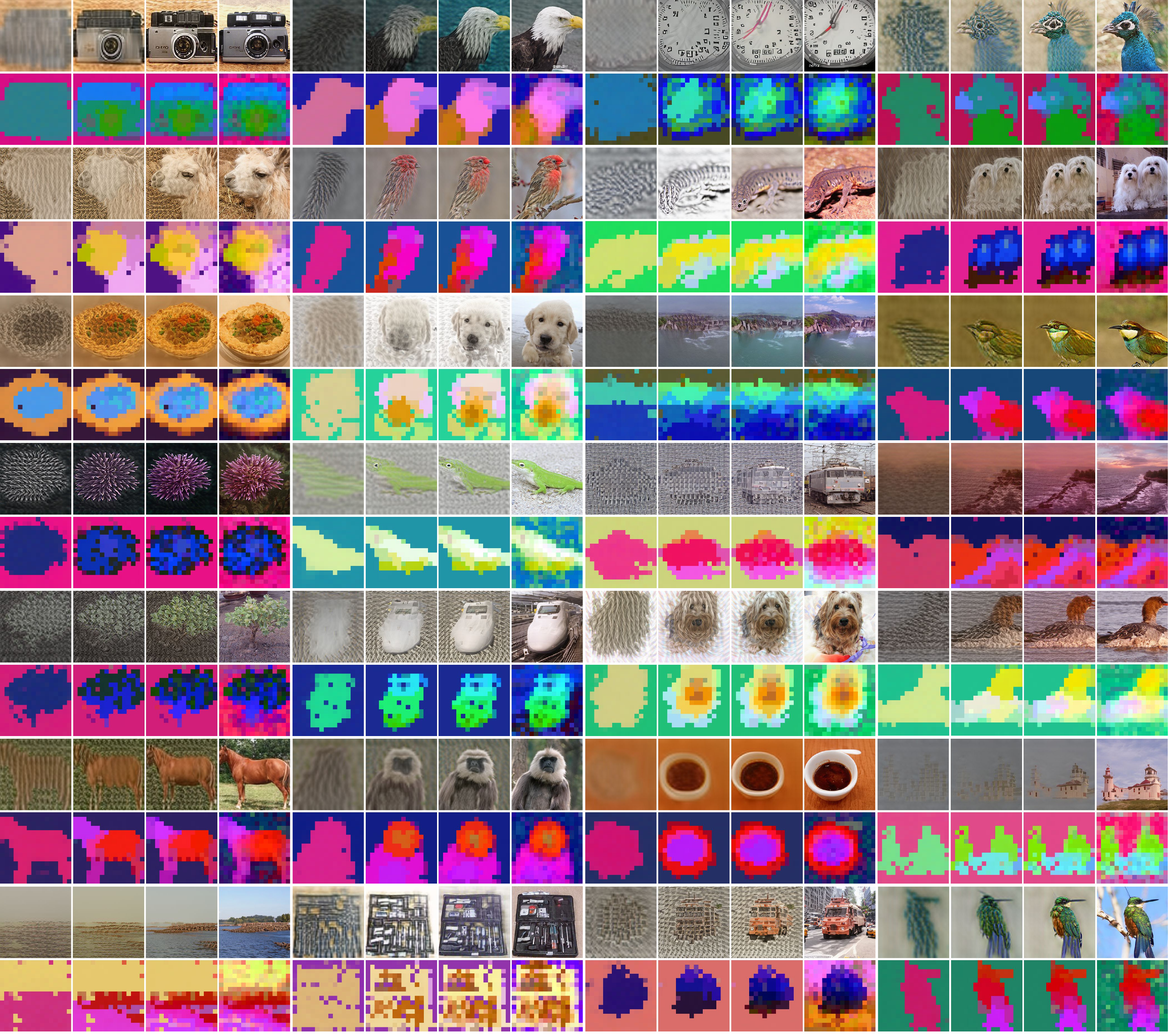}
    \vspace{-2mm}
    \captionsetup{font={stretch=0.9,footnotesize}}
    \captionof{figure}{Visualization of the full hierarchical generation process. For each sample we show the four generation levels from coarse to fine: object/background, parts, subparts, and the final latent. Latent tokens are visualized using PCA-based coloring, while the right column shows images decoded with the frozen RAE decoder. This illustrates how global structure emerges at coarse levels and progressively refines into detailed image content at the final level.}
    \label{fig:hier_only}
    \vspace{-6mm}
\end{figure*}

%% file: tables/config.tex
\begin{table}[ht]
    \centering
    \setlength{\tabcolsep}{6pt}
    \renewcommand{\arraystretch}{1.1}
    \caption{\textbf{Configuration of Trajectory Forcing.} }

    \label{tab:config-hparams}
    \vspace{-3mm}
    
    \begin{tabularx}{.9\linewidth}{X c c c}
        \noalign{\hrule height 1pt}
        \noalign{\vskip 0.1em}
        
        \textbf{configs} & \textbf{TF-B/16} & \textbf{TF-L/16} & \textbf{TF-H/16} \\
        \hline
        depth & 16 & 32 & 48 \\
        hidden dim & 768 & 1024 & 1280 \\
        attn heads & 12 & 16 & 16 \\
        noise scale & \multicolumn{3}{c}{1.0} \\
        aux-head depth & \multicolumn{3}{c}{8} \\
        class tokens & \multicolumn{3}{c}{8} \\
        time tokens & \multicolumn{3}{c}{4} \\
        level tokens & \multicolumn{3}{c}{4} \\
        linear layer init & \multicolumn{3}{c}{$\mathcal{N}(0,\sigma^2),\, \sigma^2 = 0.1/\mathrm{fan\_in}$} \\
        \hline
        epochs & 80 & 80 & 80 \\
        batch size & 1024 & 1024 & 512 \\
        precision & fp32 & bf16 & bf16 \\
        optimizer & \multicolumn{3}{c}{Muon, with $(\beta_1, \beta_2) = (0.9, 0.95)$} \\
        learning rate & \multicolumn{3}{c}{constant 0.01} \\
        ema half-life (Mimgs) & \multicolumn{3}{c}{\{500, 1000, 2000\}} \\
        \hline
        $(t, r)$ cond & \multicolumn{3}{c}{$t - r$} \\
        $t, r$ sampler & \multicolumn{3}{c}{plateau-logit-normal(0.0, 1.0)} \\
        time shift $(\alpha)$  & \multicolumn{3}{c}{10.0} \\
        \hline
        Structural loss weight & \multicolumn{3}{c}{1} \\
        Post-training FD-loss~\cite{yang2026representation} & \multicolumn{3}{c}{Inception-v3~\cite{szegedy2016rethinking}} \\
        \noalign{\hrule height 1pt}

    \end{tabularx}

\end{table}

%% file: tables/decoded_sanity_checks.tex
\begin{table}[t]
    \centering
    \setlength{\tabcolsep}{7pt}
    \renewcommand{\arraystretch}{1.15}
    \caption{\textbf{Decoded-space sanity checks in unedited regions.} We decode images before and after editing using the frozen RAE decoder and compute masked SSIM and masked LPIPS over unedited pixels only. Higher SSIM and lower LPIPS indicate that edits remain localized after decoding. Results are reported for feature edits (top) and shape edits (bottom) at editing different hierarchy levels.}
    \label{tab:decoded_sanity}
    \begin{tabular}{lccc}
        \toprule
        & \textbf{Object/Background} & \textbf{Parts} & \textbf{Subparts} \\
        \midrule
        \multicolumn{4}{l}{\textbf{Feature edits}} \\
        \cmidrule(lr){1-4}
        \cmidrule(lr){1-4}
        \hspace{2mm} Masked SSIM ($\uparrow$)  & 0.77 & 0.84 & 0.92 \\
        \hspace{2mm} Masked LPIPS ($\downarrow$) & 0.08 & 0.061 & 0.042 \\
        \midrule
        \multicolumn{4}{l}{\textbf{Shape edits}} \\
        \cmidrule(lr){1-4}
        \cmidrule(lr){1-4}
        \hspace{2mm} Masked SSIM ($\uparrow$)  & 0.89 & 0.93 & 0.96 \\
        \hspace{2mm} Masked LPIPS ($\downarrow$) & 0.028 & 0.024 & 0.013 \\
        \bottomrule
    \end{tabular}
\end{table}